\documentclass[10pt,twocolumn,letterpaper]{article}

\usepackage[pagenumbers]{cvpr} % To force page numbers, e.g. for an arXiv version

\usepackage{times}
\usepackage{epsfig}
\usepackage[utf8]{inputenc}
\usepackage{url}
\newcommand{\mycomment}[1]{}
\usepackage{graphicx}
\usepackage{amsmath}
\usepackage{amssymb}
\usepackage{subcaption}
\usepackage{caption}
\usepackage{multirow, booktabs, makecell}
\usepackage[table]{xcolor}
\usepackage[accsupp]{axessibility} 
\usepackage{colortbl}
\newcommand{\TZ}[1]{{\color{cyan}{\bf TZ: #1}}}

\usepackage[pagebackref,breaklinks,colorlinks]{hyperref}

\usepackage[capitalize]{cleveref}
\crefname{section}{Sec.}{Secs.}
\Crefname{section}{Section}{Sections}
\Crefname{table}{Table}{Tables}
\crefname{table}{Tab.}{Tabs.}

\def\cvprPaperID{9412} % *** Enter the CVPR Paper ID here
\def\confName{CVPR}
\def\confYear{2022}

\begin{document}
%\pagenumbering{gobble}
\title{MulT: An End-to-End Multitask Learning Transformer}

\author{Deblina Bhattacharjee, Tong Zhang, Sabine Süsstrunk and Mathieu Salzmann\\
School of Computer and Communication Sciences, EPFL, Switzerland\\
{\tt\small \{deblina.bhattacharjee, tong.zhang, sabine.susstrunk, mathieu.salzmann\}@epfl.ch}
}
\maketitle

%%%%%%%%% ABSTRACT
\begin{abstract}
   
 We propose an end-to-end \textbf{Mul}titask Learning \textbf{T}ransformer framework, named \textbf{MulT}, to simultaneously learn multiple high-level vision tasks, including depth estimation, semantic segmentation, reshading, surface normal estimation, 2D keypoint detection, and edge detection. Based on the Swin transformer model, our framework encodes the input image into a shared representation and makes predictions for each vision task using task-specific transformer-based decoder heads. At the heart of our approach is a shared attention mechanism modeling the dependencies across the tasks. We evaluate our model on several multitask benchmarks, showing that our MulT framework outperforms both the state-of-the art multitask convolutional neural network models and all the respective single task transformer models. Our experiments further highlight the benefits of sharing attention across all the tasks, and demonstrate that our MulT model is robust and generalizes well to new domains. Our project website is at https://ivrl.github.io/MulT/.
 %The entire model is jointly trained in an end-to-end manner with a weighted-sum loss of each task. 
 %We share the same model parameters across all task decoders instead of separately fine-tuning task-specific models, thereby reducing the number of parameters to be trained. Further, we introduce among the transformer decoders for the different tasks, revealing that attention sharing is beneficial across all tasks. 
 %Our method is evaluated on multiple benchmarks for six different vision tasks, where our MulT model outperforms not only the convolutional neural network model for the same multitask setting, but also all the respective single task transformer models, achieving an average increase of 13.2\% w.r.t. single task transformer models and an average increase of 19.5\% w.r.t. multitask convolutional neural network models. 

\end{abstract}

%%%%%%%%% BODY TEXT
\section{Introduction}
\label{sec:intro}
\begin{figure}[ht]
\centering
{\includegraphics[height=5.9cm, width=8.5cm]{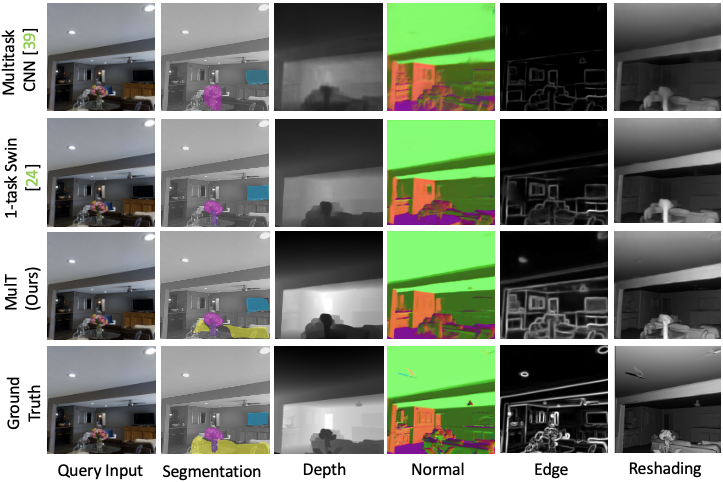}}
%\vspace{-11.3 pt}
\caption{\textbf{Motivation for MulT.} Our MulT model, which is a transformer-based encoder-decoder model with shared attention to learn task inter-dependencies, produces better results than both the dedicated 1-task transformer models (1-task Swin~\cite{swin}) and the state-of-the-art multitask CNN baseline~\cite{standley2019}. %Best seen on screen and zoomed in.  
}\label{fig:teaser}\vspace{-10pt}
\end{figure}
First proposed in~\cite{attention-is-all-you-need}, transformers have made great strides in a wide range of domains. For instance, previous works~\cite{devlin2019bert, Radford2018ImprovingLU, radford2019language, liu2019roberta, JMLR:v21:20-074, NEURIPS2019_dc6a7e65} have demonstrated that transformers trained on large datasets learn strong representations for many downstream language tasks; and models based on transformers have achieved promising results on image classification, object detection, and panoptic segmentation~\cite{pmlr-v80-parmar18a, dosovitskiy2021an, Bello_2019_ICCV, hu2018genet, ramachandran2019standalone, Wang_2018_CVPR, carion2020endtoend, zhu2021deformable, pmlr-v139-touvron21a}. In contrast to these works that focus on a single task, in this paper, we investigate the use of transformers for multitask learning.

Although a few works have studied the use of transformers to handle multiple \emph{input modalities}, such as images and text, they typically focus on a single \emph{task}, e.g., visual question
answering~\cite{hu2021unit,li2019visualbert, Lu_2020_CVPR, DBLP:journals/corr/abs-1911-06258}, with the exception of~\cite{hu2021unit}, which tackles several language tasks but a single vision one. By contrast, our goal is to connect multiple vision tasks covering the 2D, 3D, and semantic domains. To this end we address the following questions: Can a transformer model trained jointly across tasks improve the performance in each task relative to single-task transformers? Can one explicitly encode dependencies across tasks in a transformer-based framework? Can a multitask transformer generalize to unseen domains?

To the best of our knowledge, only~\cite{IPT, spatiotemporalMTL, video-multitask-transformer} have touched upon the problem of addressing multiple tasks with transformers. However, none of these works aims to encode strong dependencies across the tasks beyond the use of a shared backbone.
%used~\cite{IPT}. \MS{Did we need this citation here?}
Furthermore, IPT~\cite{IPT} handles solely low-level vision tasks, such as denoising, super-resolution and deraining, whereas~\cite{spatiotemporalMTL} focuses uniquely on the tasks of object detection and semantic segmentation and~\cite{video-multitask-transformer} on scene recognition and importance score prediction in videos. Here, we cover a much wider range of high-level vision tasks and explicitly model their dependencies.

To this end, we introduce MulT, which consists of a transformer-based encoder to transform the input image into a latent representation shared by the tasks, and transformer decoders with task-specific 
heads producing the final predictions for each of the tasks. While the MulT encoder mainly utilizes the self-attention mechanism~\cite{bahdanau2016neural, parikh-etal-2016-decomposable} to extract intrinsic features, as most transformers, we equip the decoders with a shared attention mechanism across the different vision tasks, thus allowing the overall framework to encode task dependencies.
Thus, we leverage the query and key vectors from the encoder along with the task-specific values in the decoder to predict the task-specific outputs. 
Our contributions can be summarized as follows:
\begin{itemize}
    \item We propose an end-to-end multitask transformer architecture that handles multiple high-level vision tasks in a single model. 
    
    \item We introduce a shared attention between the transformer decoders of the multiple tasks. This shared attention mechanism further improves the performance of each vision task.
    
    \item Our framework lets us learn the inter-dependencies across high-level vision tasks.

    \item We show that our model generalizes and adapts to new domains with a lower average error on the different vision tasks than the existing multitask convolutional models~\cite{zamir2020consistency, standley2019}. 
    %\MS{Any difference between new and unseen?} \DB{no difference: new domain would be blurred images of Taskonomy, and unseen domain would be a completely different dataset- but overall it means the same thing. I've removed unseen.}  \MS{Lower than what?}\DB{corrected}.
\end{itemize}

Our exhaustive experiments and analyses across a variety of tasks show our MulT model not only improves the performance over single-task architectures, but also outperforms the state-of-the-art multitask CNN-based models (as shown in Figure~\ref{fig:teaser}) on standard benchmarks, such as Taskonomy~\cite{taskonomy2018}, Replica~\cite{replica}, NYU~\cite{NYU} and CocoDoom~\cite{cocodoom} .

\begin{figure*}[ht]
\centering
{\includegraphics[ width=0.90\linewidth ]{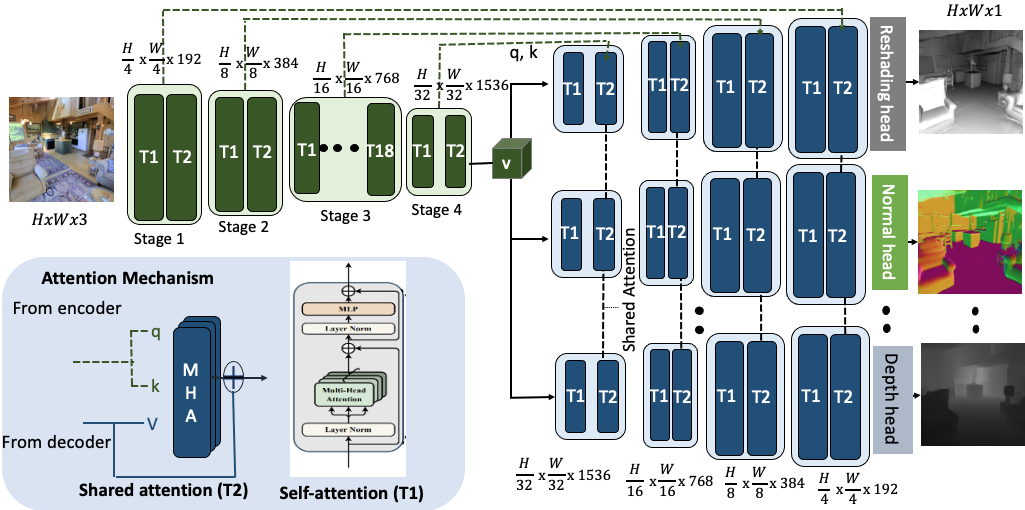}}
%\vspace{-11.3 pt}
\caption{{\textbf{Detailed overview of our MulT architecture.} Our MulT model builds upon the Swin transformer~\cite{swin} backbone and models the dependencies between multiple vision tasks via a shared attention mechanism (shown in the bottom left), which we introduce in this work. The encoder module (in green) embeds a shared representation of the input image, which is then decoded by the transformer decoders (in blue) for the respective tasks. Note that the transformer decoders have the same architecture but different task heads. The overall model is jointly trained  in a supervised manner using a weighted loss~\cite{gradnorm} of all the tasks involved. For clarity, only three tasks are depicted here. } 
}\label{fig:detailed-model}\vspace{-10pt}
\end{figure*}
%-------------------------------------------------------------------------
\section{Related Work}
\paragraph{Multitasking.}
In its most conventional form,
multi-task learning predicts multiple outputs out of a shared encoder/representation for an input~\cite{zhang2021survey}. Prior works~\cite{taskonomy2018, zamir2020consistency, standley2019, strezoski2019taskrouting, endtoendMTL} follow this architecture to jointly learn multiple vision tasks using a CNN. Leveraging this encoder-decoder architecture, IPT~\cite{IPT} was the first transformer-based multitask network aiming to solve low-level vision tasks after fine-tuning a large pre-trained network. This was followed by~\cite{spatiotemporalMTL}, which jointly addressed the tasks of object detection and semantic segmentation. Recently, \cite{video-multitask-transformer} used a similar architecture for scene and action understanding and score prediction in videos. However, none of these works connect such a wide range of vision tasks as we do, including 2D, 3D, and semantic domains. Furthermore, they do not explicitly model the dependencies between the tasks, which we achieve via our shared attention mechanism.
\vspace{-10pt}
\paragraph{Transformers.}
Transformers~\cite{attention-is-all-you-need} were originally introduced for language tasks, in particular for machine translation where they showed impressive improvements over recurrent-based encoder-decoder
architectures. Since then they have been widely applied to a great range of problems, including speech recognition~\cite{gulati2020conformer} and language modeling~\cite{dai2019transformerxl, devlin2019bert}.
In the vision domain, transformers have been used to extract visual features, replacing CNNs for object detection, image classification, segmentation and video representation learning~\cite{carion2020endtoend, zhu2021deformable, pmlr-v80-parmar18a, dosovitskiy2021an, Bello_2019_ICCV, hu2018genet, mttransunet}. Recently, several works, such as UniT~\cite{hu2021unit} and VILBERT-MT~\cite{li2019visualbert}, have learned multiple tasks from multimodal domains, such as vision and text. Here, however, we focus on a single input modality: images.
%However, with the exception of~\cite{IPT, spatiotemporalMTL, video-multitask-transformer}, none of these works tackle multitask learning, which is the problem we aim to solve. Moreover, we aim to handle multiple tasks in the 2D, 3D and semantic domains. 
\vspace{-10pt}
\paragraph{Learning task inter-dependencies.}
Taskonomy~\cite{taskonomy2018} studied the relationships between multiple visual tasks for transfer learning and introduced a dataset with 4 million images and corresponding labels for 26 tasks. Following this, a number of recent works have further studied tasks relationships for transfer learning~\cite{NEURIPS2019_f490c742, DBLP:journals/corr/abs-1903-01092, dwivedi2019representation, achille2019task2vec}. However, these works differ from multitask learning, in the sense that they analyze a network trained on a source task and applied to a different target task, whereas we study the effect of leveraging multiple tasks during training. In~\cite{standley2019}, Standley et al. found notable differences between transfer task affinity and multi-task affinity and showed the benefits of leveraging structural similarities between tasks at all levels for multitask learning. In this work, we further study the task inter-dependencies, but by designing a multitask transformer model instead of a CNN one. Our MulT model lets us learn the inter-dependencies across high-level vision tasks and further improves the task inter-dependencies seen in CNN-based models.
\vspace{-10pt}
\paragraph{Attention mechanisms.}
While there have been a myriad of attention mechanisms~\cite{chu2021Twins, wang2021pvtv2, xu2021coscale, yang2021focal, wang2021crossformer, chen2021regionvit} to exploit long range dependencies using transformers, none of the prior works utilize a cross-task shared attention for multitask learning. This is what we propose in this work to handle multiple vision tasks.
%, that in turn learns the task inter-dependencies, to report improved performances on the respective tasks w.r.t. 1-task dedicated networks. 

%-------------------------------------------------------------------------

\section{MulT: A Multitask Transformer}
Our model, MulT, follows the principle of a transformer encoder-decoder architecture~\cite{attention-is-all-you-need}. It consists of a transformer-based encoder to map the input image to a latent representation shared by the tasks, followed by transformer decoders with task-specific heads producing the predictions for the respective tasks. Figure~\ref{fig:detailed-model} shows an
overview of our MulT framework. 
%We consider multiple vision tasks in our framework. 
For our transformer-based encoder, we use a pyramidal backbone, named the Swin Transformer~\cite{swin}
to embed the visual features into a list of hidden states that incorporates global contextual information. We then apply the transformer decoders to progressively decode and upsample the tokenized maps from the encoded image. Finally, the representation from the transformer decoder is passed to a task-specific head, such as a simple two layer classifier (in the case of segmentation), which outputs the final predictions.  Given the simplicity of MulT, it can be extended easily to more tasks. We empirically show that our model can jointly learn 6
different tasks and generalizes well to new domains. The following sections
describe the details of each component in MulT.

\subsection{Encoder Module}
%We use a pyramidal backbone, named the Swin Transformer~\cite{swin} operating exclusively on low-resolution features, which is able to provide multi-scale information to the decoders. This has significant advantage over a columnar transformer encoder like~\cite{dosovitskiy2021an}, as Swin is capable of processing feature embeddings at gradually decreasing resolutions, allowing to retain more fine-grained details in the early stages of the encoding process. 
For the encoder, we adopt Swin-L~\cite{swin}, which applies stacked transformers to features of gradually decreasing resolution in a pyramidal manner, hence producing hierarchical multi-scale encoded features, as shown in Figure~\ref{fig:detailed-model}. In particular, following the ResNet~\cite{resnet} structure and design rules, four stages are defined in succession: each of them contains a patch embedding step, which reduces the spatial resolution and increases the channel dimension, and a columnar sequence of transformer blocks. The initial basic patch embedding in the first stage is performed with square patches of size $p_H = p_W = 4$ and with channel size C = 192, without the addition of the ‘class’ token; the patch merging in all three subsequent stages takes the output tokens of the previous stage, reshapes them in a 2D representation and aggregates neighboring tokens in non-overlapping patches of size $p_H = p_W = 2$ through channel-wise concatenation and a linear transformation that halves the resulting number of channels (hence doubles the number of channels with respect to the input tokens). This approach halves the resolution and doubles the channel dimension at every intermediate stage, matching the behavior of typical fully-convolutional backbones and producing a feature pyramid (with output sizes of {1/4, 1/8, 1/16, 1/32} of the original resolution) compatible with most previous architectures for vision tasks. 

Following~\cite{resnet}, most of the computation is concentrated in the third stage: Out of a total of $N = 24$ transformer encoders, 2 blocks are in the first, second and fourth stage and 18 are in the third stage. In each block, the self-attention is repeated according to the number of heads used and depending on the stage of the encoding process. This is done to match the increase in the channel dimensions, where the dimensions $M = \{6, 12, 24, 48\}$ in the first, second, third and fourth stage, respectively. However, the high resolution in the first two stages does not allow the use of global self-attention, due to its quadratic complexity with respect to the token sequence length. To solve this issue, in all stages, the tokens, that are reshaped in a 2D representation, are divided into non-overlapping square windows of size $h = w = 7$, and the intra-window self-attention is independently computed for each of them. This means that each token attends to only the tokens in its own window, both as a query and as a key/value. A possible downside of this approach could be that the restriction to fixed local windows completely stops any type of global or long-range interaction. The adopted solution is to alternate regular window partitioning with another non-overlapping partitioning in which the windows are shifted by half their size, $\lfloor h/2 \rfloor = \lfloor w/2 \rfloor = 3$, both in the height and width dimensions. This has the effect of gradually increasing the virtual receptive field of the subsequent attention computations. 

%The advantage of this self-attention mechanism w.r.t. the previous transformer encoders~\cite{dosovitskiy2021an} is that all query pixels in a window share the same key set, which allows efficient memory access and low latency, while still maintaining comparable expressiveness; furthermore, the locality characteristic of this type of attention mechanisms does not negatively impact performance due to the high correlation present in visual inputs.
\subsection{Decoder Module}
Inspired by the two CNN-based decoders proposed in~\cite{SETR}, we develop corresponding conceptually similar transformer-based versions. The general idea is to replace convolutional layers with windowed transformer blocks. Specifically, our decoder architecture consists of four stages, each containing a sequence of 2 transformer blocks for a total of 8. In each stage, the two sequential transformer blocks allow us to leverage inter-window connectivity by alternating regular and shifted window configurations as in the encoder.  Between consecutive stages, we use an upsampling layer to double the spatial resolution and half the channel dimension; we therefore adjust the number of attention heads accordingly to {48, 24, 12, 6}, in the first, second, third and fourth stage, respectively. The spatial/channel shape of the resulting feature maps matches the outputs of the encoder stages, which are delivered to the corresponding decoder stages by skip connections. This yields an hourglass structure with mirrored encoder-decoder communication: the lower-resolution stages of the decoder are guided by the higher-level deeper encoded features and the higher-resolutions stages of the decoder are guided by the lower-level shallower encoded features, allowing to gradually recover information in a coarse to fine manner and to exploit the different semantic levels where they are more relevant. Note that the first transformer block in each stage of the decoder uses a regular window partitioning while the second uses a shifted window partitioning; this can easily be extended to using a longer sequence of transformer blocks, as long as the length is a multiple of 2, which makes it possible to alternate between the two configurations. 

To perform multitask prediction, we share the encoder across all tasks and use task-specific decoders with the same architecture but different parameter values. We then simply 
%instantiate the decoder with separate task-specific heads
append task-specific heads to the decoder. For instance, a model jointly trained for semantic segmentation and depth prediction will have two task-specific heads: one predicting $K$ channels followed by a softmax for semantic segmentation and one predicting a single channel followed by a sigmoid for depth estimation.

\subsection{Shared Attention}
To account for the task dependencies beyond sharing encoder parameters, we develop a shared attention mechanism that integrates the information contained in the encoded features into the decoding stream.
%as well as the task inter-dependencies, while the overall structure of the windowed transformer block in the decoders and the other components are unchanged. 
Let us now describe how this mechanism works for one particular decoder stage. Note that we apply the same procedure for all decoder stages.

\begin{figure}[t]
\centering
{\includegraphics[height=6cm, width=8cm ]{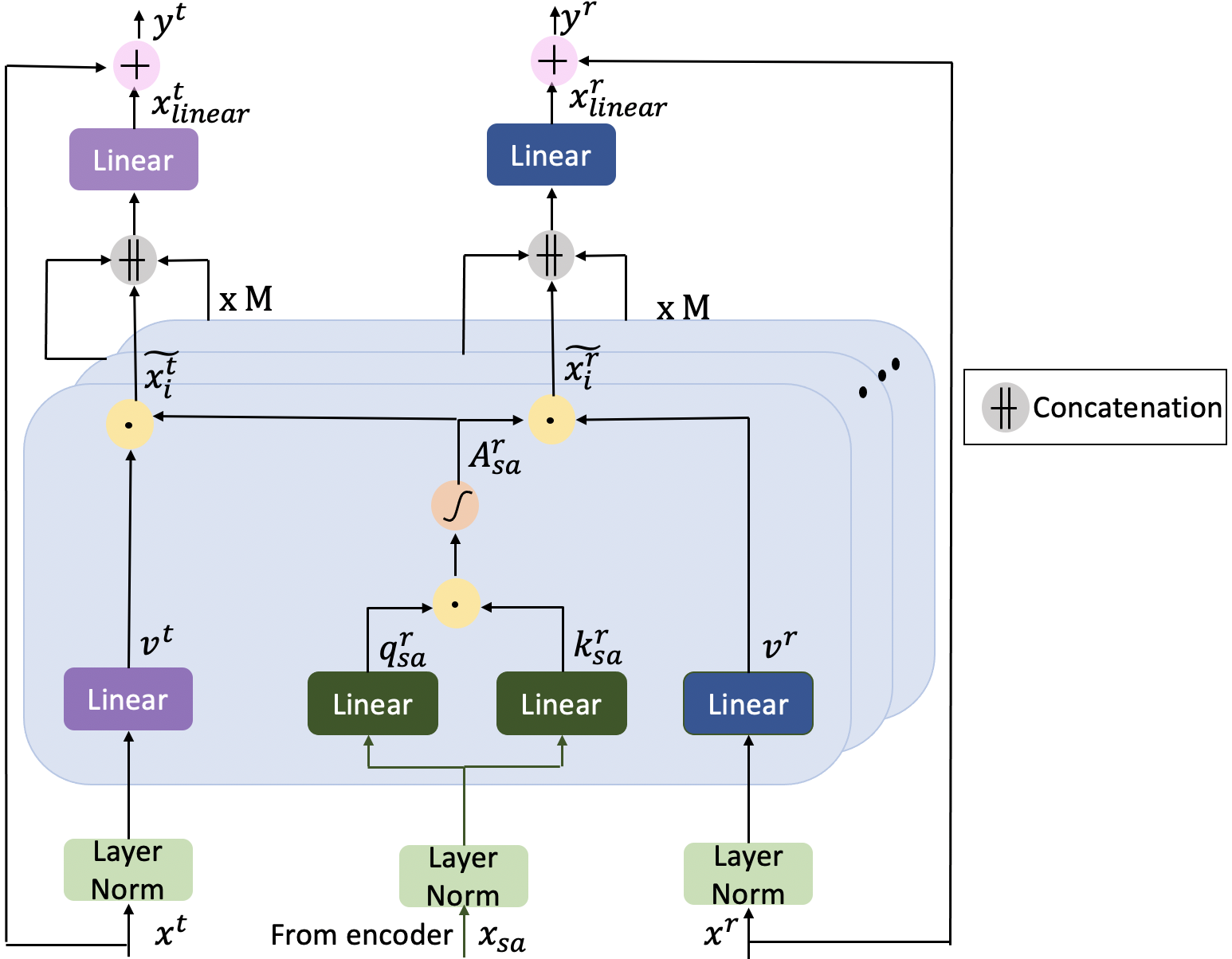}}
%\vspace{-11.3 pt}
\caption{Overview of our \textbf{shared attention} mechanism. 
%\TZ{Maybe use LN to replace Norm, and make the font larger}\DB{Done}
}\label{fig:shared-attention-detailed}\vspace{-10pt}
\end{figure}

Formally, for one task $t$ and one particular decoder stage, let $x^t$ denote the upsampled output of the previous stage, and $x_{sa}$ the output of the encoding stage operating at the same resolution. As illustrated in Figure~\ref{fig:shared-attention-detailed}, the decoder stage takes both  $x^t$ and $x_{sa}$ as input. The standard way to compute self-attention for task $t$ would be to obtain the key, query and value vectors from its own decoder output $x^t$ only. By contrast, for our shared attention, we use only one of the task streams to calculate the attention. That is, we compute a query $q_{sa}^{r}$ and a key $k_{sa}^r$ from $x_{sa}$ (coming from the encoder) by using the linear layers, shown in Figure~\ref{fig:shared-attention-detailed}, of the decoder of one particular reference task $r$. To nonetheless reflect the fact that the output of the decoder for task $t$ should be related to this particular task, we compute the values $v^t$ using the previous stage output $x^t$ for task $t$. Thus, we compute attention values from the reference task $r$ as
\begin{equation} \label{eq:method_shared_attention}
\begin{aligned}
    A^r_{sa}=\text{softmax}\left(\frac{q^r_{sa}.{k^r_{sa}}^T}{\sqrt{C^r_{qkv}}}+B^r\right),\\
\end{aligned}                   
\end{equation}
where $C^r$ is the number of channels and $B^r$ is the bias. For any task $t$, we then compute $\tilde{x}^t=A^r_{sa}v^t.$ This $\tilde{x}^t$ is then used by the self-attention head \text{head}$^t_i(.,.)$ to compute \text{head}$^t_i(\tilde{x}^t_i,{W^t_i})= \tilde{x}^t_i\cdot {W^t_i}$, where ${W^t_i}$ is the learnt attention weight for task $t$ and $\tilde{x}^t_i$ is the \text{i}$^{th}$ channel, respectively. Note that this formulation represents the $i^{th}$ instance of the self-attention, which is repeated $M$ times to obtain a multihead attention as \text{MHA}$^t(.,.)$ for task $t$. 
%\TZ{another thing is SA-head, since attention has already calculated by $A{_sa}^r$, maybe we just need to call it $head_1$}yes
Following which, we compute $x^t_{linear}$ by linearly projecting the output of \text{MHA}$^t(.,.)$. Finally, we obtain $y^t$ as follows:
\begin{equation} \label{eq:method_shared_attention-MHA}
\begin{aligned}
    &\text{MHA}^t(\tilde{x}^t,W)=\text{Concat}(\text{head}^t_1,\cdots, \text{head}^t_M)\mathbf{W}\;,\\
    & x^t_{linear}= \text{MHA}^t(.,.)\;,\\
    & y^t = x^t + x^t_{linear}\;,
\end{aligned}
\end{equation}
where $\mathbf{W}$ indicates the multi-head attention weight. Empirically, we have found that the attention from the surface normal task stream benefits our 6-task MulT model, and we thus take this task as reference task $r$, whose attention is shared across the tasks. As shown in Figure~\ref{fig:shared-attention-detailed}, $x^r$ is the upsampled output of the previous stage of a particular decoder for the reference task, taken here as surface normal prediction.

\begin{figure}[t]
\centering
{\includegraphics[ height= 4.8cm, width=8.5cm ]{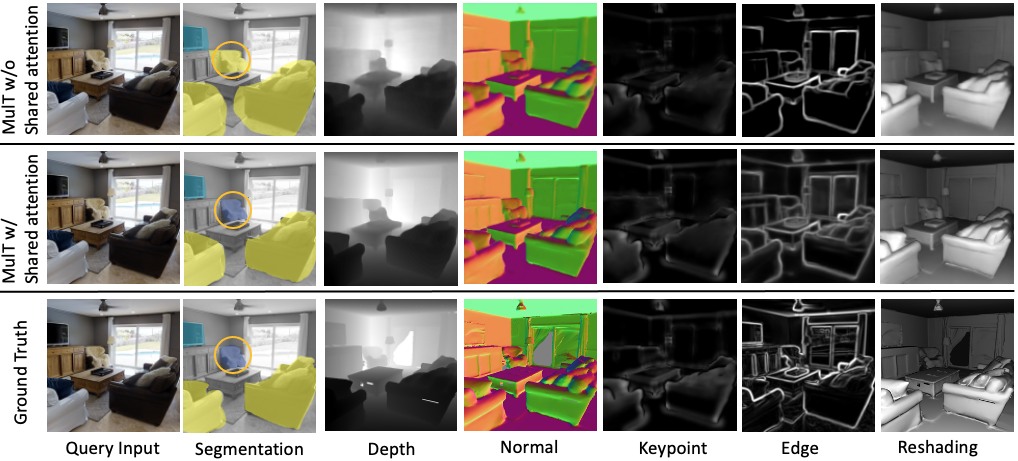}}
%\vspace{-11.3 pt}
\caption{{\textbf{Motivation of the shared attention mechanism on our MulT model.} The shared attention mechanism learns the task inter-dependencies and improves the prediction for each task. For instance, the yellow circled region shows how our MulT model with shared attention across tasks improves the semantic segmentation performance, where the chair mask is correctly classified in our predictions as in the ground truth. However in the MulT model without the shared attention, the chair is predicted as a couch mask. Best viewed on screen and when zoomed in.}
}\label{fig:effect-of-shared-attention}\vspace{-10pt}
\end{figure}

Note that our shared attention differs from the co-attention introduced in prior works\cite{Chefer_2021_CVPR}, where the value and key are passed via a skip connection from the encoder layers. Figure~\ref{fig:effect-of-shared-attention} shows the effect of adding our shared attention mechanism across the tasks, where our MulT with the shared attention mechanism improves the results across all the tasks in comparison with our MulT model without the shared attention. 
%Specifically, we can see that we achieve results on par with the ground truth for the task of semantic segmentation.

\vspace{-15pt}
\paragraph{Task Heads and Loss.} The feature maps from the transformer decoder modules are input to different task-specific heads to make subsequent predictions. Each class head includes a single linear layer to output a $H \times W \times 1$ map, where $H$, $W$ are the input image dimensions. We employ a weighted sum~\cite{gradnorm} based task-specific losses to jointly train the network, where the losses are calculated between the ground truth and final predictions for each task. In particular, we use cross-entropy for segmentation, rotate loss~\cite{taskonomy2018} for depth, and $L1$ loss for surface normals, 2D keypoints, 2D edges and reshading, respectively. Note that we employ these losses to maintain consistency with the baselines~\cite{taskonomy2018, standley2019, zamir2020consistency}.

\section{Experiments and Results}
To provide a thorough analysis of MulT and also compare it with well-established prior work, we
experiment with jointly learning prominent, high-level vision tasks. 

\subsection{Datasets}
We evaluate MulT using the following datasets:
     \vspace{-13pt}
     \paragraph{Taskonomy~\cite{taskonomy2018}} is used as our main training dataset. It comprises 4 million real images of indoor scenes with multi-task annotations for each image. The experiments were performed using the following 6 tasks: \textit{semantic segmentation $\mathcal{(S)}$,  depth (zbuffer) $\mathcal{(D)}$, surface normals $\mathcal{(N)}$, 2D keypoints $\mathcal{(K)}$, 2D (Sobel) texture edges (E) and reshading $\mathcal{(R)}$}. The tasks were selected to cover 2D, 3D, and semantic domains and have sensor-based/semantic ground truth. We report results on the Taskonomy test set. 
     
     \vspace{-13pt}
     \paragraph{Replica~\cite{replica}} comprises high-resolution 3D ground truth and enables more reliable evaluations of fine-grained details. We test all the networks on 1227 images from Replica (with and without fine-tuning).
     \vspace{-13pt}
     
    \paragraph{NYU~\cite{NYU}} comprises 1449 images from 464 different indoor scenes. We test all the networks on NYU (with and without fine-tuning). 
    \vspace{-13pt}
    
     \paragraph{CocoDoom~\cite{cocodoom}} contains synthetic images from the Doom video game. We use it as an out-of-training-distribution dataset.

\subsection{Training Details}
We jointly train MulT on multiple tasks, including, semantic segmentation, depth estimation, 2D keypoint detection, 2D edge detection, surface normal estimation and reshading. In our implementation, we train with a
batch size of 32 on 32 Nvidia V100-SXM2-32GB
GPUs in a distributed fashion, using PyTorch.
We use the weighted Adam optimizer~\cite{adam-w} with a
learning rate of 5e-5 and the warm-up cosine learning rate schedule (using 2000 warm-up iterations). The optimizer updates the model parameters based on gradients from the task losses. 
\subsection{Baselines}
We compare our MulT model with the following state-of-the-art baselines.
\vspace{-15pt}
\paragraph{Baseline UNet (for single-task or independent learning)} constitutes our CNN-based baseline. We use it as a reference for all the multitask models.
\vspace{-15pt}
    %All CNN based multitask models, transfer learning model and our MulT model are compared against this single-task Unet baseline. We report the relative performance of the respective models compared with the single-task Unet baseline in Table~\ref{tb:MulT-sixstream-results-taskonomy} and Table~\ref{tb:MulT-sixstream-results-replica-nyu}.
\paragraph{Baseline Swin transformer~\cite{swin} (for single-task or independent learning)} constitutes the single task transformer baseline. It is almost identical to our MulT model, except for not including shared attention and for being trained with only one dedicated task. We use it to evaluate the benefits of our multitask learning strategy
    %MulT model relative to the single-task Swin baseline in Table~\ref{tb:MulT-sixstream-results-taskonomy} and Table~\ref{tb:MulT-sixstream-results-replica-nyu}.
\vspace{-15pt}
\paragraph{Multi-task learning~\cite{kokkinos2016ubernet} (MTL)} comprises a network with one shared encoder and multiple decoders each dedicated to a task. This baseline further identifies if tasks are inter-dependent, such that a shared representation can give comparable performance across multiple tasks, without explicitly adding task constraints. 
\vspace{-13pt}
\paragraph{Taskonomy~\cite{taskonomy2018}} studies the relationships between multiple visual tasks for transfer learning. 
\vspace{-13pt}
\paragraph{Taskgrouping~\cite{standley2019}} studies task compatibility in multitask learning, thus providing a framework for determining which tasks should be trained jointly and which tasks should be trained separately.
\vspace{-13pt}
\paragraph{Cross-task consistency~\cite{zamir2020consistency}} presents a general and data-driven framework for augmenting standard supervised learning with cross-task consistency. It is inspired from Taskonomy~\cite{taskonomy2018} but adds a consistency constraint to learn multiple tasks jointly.

Note that we do not compare our method with the contemporary work~\cite{hu2021unit} as it focuses on \emph{bimodal} multitask learning for vision- and language-related tasks. By contrast, in this work, we tackle \emph{unimodal} multitask learning for high-level vision tasks. All the multitask baselines were trained using their best model configurations as in~\cite{kokkinos2016ubernet, taskonomy2018, standley2019, zamir2020consistency}, respectively.
%dense prediction tasks as well as vision tasks like edge and keypoint detection.
\begin{table}[ht]
\setlength\tabcolsep{3pt}
\centering
\scalebox{0.85}{
\arrayrulecolor{black}
\begin{tabular}{!{\color{white}\vrule}l!{\color{white}\vrule}c!{\color{white}\vrule}c!{\color{white}\vrule}c!{\color{white}\vrule}c!{\color{white}\vrule}c!{\color{white}\vrule}c}
\hline
\multicolumn{7}{l}{~ ~ ~ ~ ~ ~ ~ ~ ~ ~ ~ ~~\textbf{ ~ Relative Performance On}}   \\
                                                    & $\mathcal{S}$     & $\mathcal{D}$ & $\mathcal{N}$ & $\mathcal{K}$ & \textit{E} & $\mathcal{R}$  \\ 
\arrayrulecolor{black}\hline
$\mathcal{S}$                                              & -                                                & +3.83\%                                           & -1.42\%                                            & -1.33\%                                              & +33.9\%                                         & -0.80\%                                              \\ 
$\mathcal{D}$                                               & +4.83\%                                          & -                                                 & +2.77\%                                            & -1.92\%                                              & +35.2\%                                         & +3.93\%                                              \\ 
$\mathcal{N}$                                             & +11.3\%                                          & +8.35\%                                           & -                                                  & +91.2\%                                              & +77.1\%                                         & +9.09\%                                              \\ 
$\mathcal{K}$                                            & +5.11\%                                          & +0.57\%                                           & -6.88\%                                            & -                                                    & +70.1\%                                         & +0.21\%                                              \\ 
\textit{E}                                               & +6.09\%                                          & +4.33\%                                           & -0.73\%                                            & +4.75\%                                              & -                                               &       +5.11\%                                               \\ 
$\mathcal{R}$ & +8.61\%                                          & +4.45\%                                           & +5.91\%                                            & +1.95\%                                              & +33.9\%                                         & -                                                    \\
\arrayrulecolor{black}\hline
\end{tabular}}%
\setlength{\abovecaptionskip}{1mm}
\caption{\textbf{Quantitative comparison of our MulT model with a single-task dedicated Swin transformer baseline~\cite{swin}.} Our MulT model is jointly trained in a pairwise manner on the Taskonomy benchmark~\cite{taskonomy2018}. For instance, in the first row, second column, we show the results of our MulT model trained with semantic segmentation and depth in a pairwise manner, and tested on the task of depth estimation. The relative performance percentage for each task is evaluated by taking the percentage increase or decrease w.r.t. the single-task baseline. The results here are reported on the Taskonomy test set. The columns show the task tested on, and the rows show the other task used for
training.}
   \label{tb:MulT-pairwise-results}%
\vspace{-10pt}
\end{table}%

\begin{table*}[ht]
\setlength\tabcolsep{3pt}
\centering
\scalebox{0.85}{
\arrayrulecolor{black}
\begin{tabular}{!{\color{white}\vrule}l!{\color{white}\vrule}c!{\color{white}\vrule}c!{\color{white}\vrule}c!{\color{white}\vrule}c!{\color{white}\vrule}c!{\color{white}\vrule}c}
\hline
\multicolumn{7}{l}{~ ~ ~ ~ ~ ~ ~ ~ ~ ~ ~ ~ ~ ~ ~ ~ ~ ~ ~ ~ ~ ~ ~ ~ ~ ~ ~ ~ ~ ~ ~ \textbf{ ~ Relative Performance On}}   \\
\hline
\multicolumn{7}{l}{~ ~ ~ ~ ~ ~ ~ ~ ~ ~ ~ ~ ~ ~ ~ ~ ~ ~ ~ ~ ~ ~ ~ ~ ~ ~ ~ ~ ~ ~ ~ \textbf{ ~ Taskonomy Test Set~\cite{taskonomy2018}}}   \\
                                                    & $\mathcal{S}$     & $\mathcal{D}$ & $\mathcal{N}$ & $\mathcal{K}$ & \textit{E} & $\mathcal{R}$  \\ 
\arrayrulecolor{black}\hline
MTL~\cite{kokkinos2016ubernet} vs 1-task CNN~\cite{unet}                                             & +2.05\%                                                & +3.11\%                                           & +4.38\%                                            & -1.29 \%                                              & +45.22\%                                         & +2.99\%                                              \\ 
Taskonomy~\cite{taskonomy2018} vs 1-task CNN~\cite{unet}                                               & +2.63\%                                          & -3.82 \%                                              & +2.95\%                                            & +10.13 \%                                            & +59.05\%                                        & +4.52\%                                             \\ 
Taskgrouping~\cite{standley2019} vs 1-task CNN~\cite{unet}                                           & +6.24\%                                         & +3.36\%                                          & +4.23\%                                                 & +21.77\%                                             & +73.6 \%                                         & +5.79\%                                              \\ 
Cross-task~\cite{zamir2020consistency} vs 1-task CNN~\cite{unet}                                           & +9.01\%                                         & +6.77\%                                           & +5.61\%                                            & +23.20\%                                                    &+75.8 \%                                         & +11.1\%                                              \\ 
\hline
\textbf{MulT} vs 1-task Swin~\cite{swin}   & \underline{+19.7\%}                                          & \underline{+10.2\%}                                           & \underline{+8.72\%}                                            & \underline{+94.75\%}                                              & \underline{+88.8\%}                                              & \underline{+16.4\%}                                               \\ 
\textbf{MulT} vs 1-task CNN~\cite{unet} & \textbf{+21.6\%}                                          & \textbf{+11.5\%}                                           & \textbf{+9.71\%}                                            & \textbf{+97.04\%}                                              & \textbf{+92.9\%}                                         & \textbf{+21.0\%}                                                   \\
\arrayrulecolor{black}\hline
\end{tabular}}%
\setlength{\abovecaptionskip}{1mm}
\caption{\textbf{Quantitative comparison of our MulT model with baselines when jointly trained for six tasks on the Taskonomy benchmark~\cite{taskonomy2018}.} Our six-task MulT model consistently outperforms all the baselines, including the multitasking CNN baselines and the single-task CNN and Swin baselines. The relative performance percentage for each task is evaluated by taking the percentage increase or decrease w.r.t. the single-task baseline. The results here are reported on the Taskonomy test set. Bold and underlined values show the best and second-best results, respectively.  } 
   \label{tb:MulT-sixstream-results-taskonomy}%
\vspace{-10pt}
\end{table*}%

\begin{table*}[ht]
\setlength\tabcolsep{3pt}
\centering
\scalebox{0.8}{
\arrayrulecolor{black}
\begin{tabular}{!{\color{white}\vrule}l!{\color{white}\vrule}c!{\color{white}\vrule}c!{\color{white}\vrule}c!{\color{black}\vrule}c!{\color{white}\vrule}c}
\hline
\multicolumn{6}{l}{~ ~ ~ ~ ~ ~ ~ ~ ~ ~ ~ ~ ~ ~ ~ ~ ~ ~ ~ ~ ~ ~ ~ ~ ~ ~ ~ ~ ~ ~ ~ \textbf{ ~ Relative Performance On}}   \\ \hline
                                                    & \multicolumn{3}{l}{~ ~ ~ ~ ~ \textbf{ ~ Replica Dataset~\cite{replica}}} & \multicolumn{2}{l}{~ ~ \textbf{ ~ NYU Dataset~\cite{NYU}}} \\ 
                                                    & $\mathcal{D}$     & $\mathcal{N}$ & $\mathcal{R}$ & $\mathcal{S}$  & $\mathcal{D}$  \\ 
\arrayrulecolor{black}\hline
MTL~\cite{kokkinos2016ubernet} vs 1-task CNN~\cite{unet}                                             & +2.53\%                                                &+3.03\%                                           & +1.87\%                                            &+1.13\%                                              & +2.72\%                                                                                      \\ 
Taskonomy~\cite{taskonomy2018} vs 1-task CNN~\cite{unet}                                               &-4.55\%                                          &+1.99\%                                              &+3.33\%                                            &  +2.05\%                                            &-4.07\%                                                                                  \\ 
Taskgrouping~\cite{standley2019} vs 1-task CNN~\cite{unet}                                           & +2.75\%                                         & +4.09\%                                          & +5.47\%                                                 & +6.01\%                                             &  +2.91\%                                                                                     \\ 
Cross-task~\cite{zamir2020consistency} vs 1-task CNN~\cite{unet}                                           & +5.10\%                                         & +4.33\%                                           & +9.55\%                                            & +8.10\%                                                    & +5.71\%                                                                                   \\ 
\hline
\textbf{MulT} vs 1-task Swin~\cite{swin}   & \underline{+8.33\%}                                          & \underline{+7.05\%}                                           & \underline{+14.2\%}                                            & \underline{+13.3\%}                                              & \underline{+8.54\%}                                                                                            \\ 
\textbf{MulT} vs 1-task CNN~\cite{unet} & \textbf{+10.1\%}                                          & \textbf{+8.59\%}                                           & \textbf{+19.6\%}                                            & \textbf{+15.7\%}                                              & \textbf{+10.4\%}                                                                                          \\
\arrayrulecolor{black}\hline
\end{tabular}}%
\setlength{\abovecaptionskip}{1mm}
\caption{\textbf{Quantitative comparison of our MulT model with baselines on the Replica benchmark and the NYU benchmark.} We apply our MulT model, jointly trained on 6 tasks on the Taskonomy dataset, to test the depth, normals and reshading prediction performances on the Replica dataset~\cite{replica}, and the segmentation and depth prediction performance on the NYU dataset~\cite{NYU}. Our six-task MulT model consistently outperforms all the baselines, including the multitasking CNN baselines and the single-task CNN and Swin baselines. The relative performance percentage for each task is evaluated by taking the percentage increase or decrease w.r.t. the single-task baseline. Bold and underlined values show the best and second-best results, respectively.  } 
   \label{tb:MulT-sixstream-results-replica-nyu}%
\vspace{-10pt}
\end{table*}%
\subsection{Quantitative Results}
\begin{figure*}[ht]
\centering
{\includegraphics[ height= 8cm, width=13.3cm]{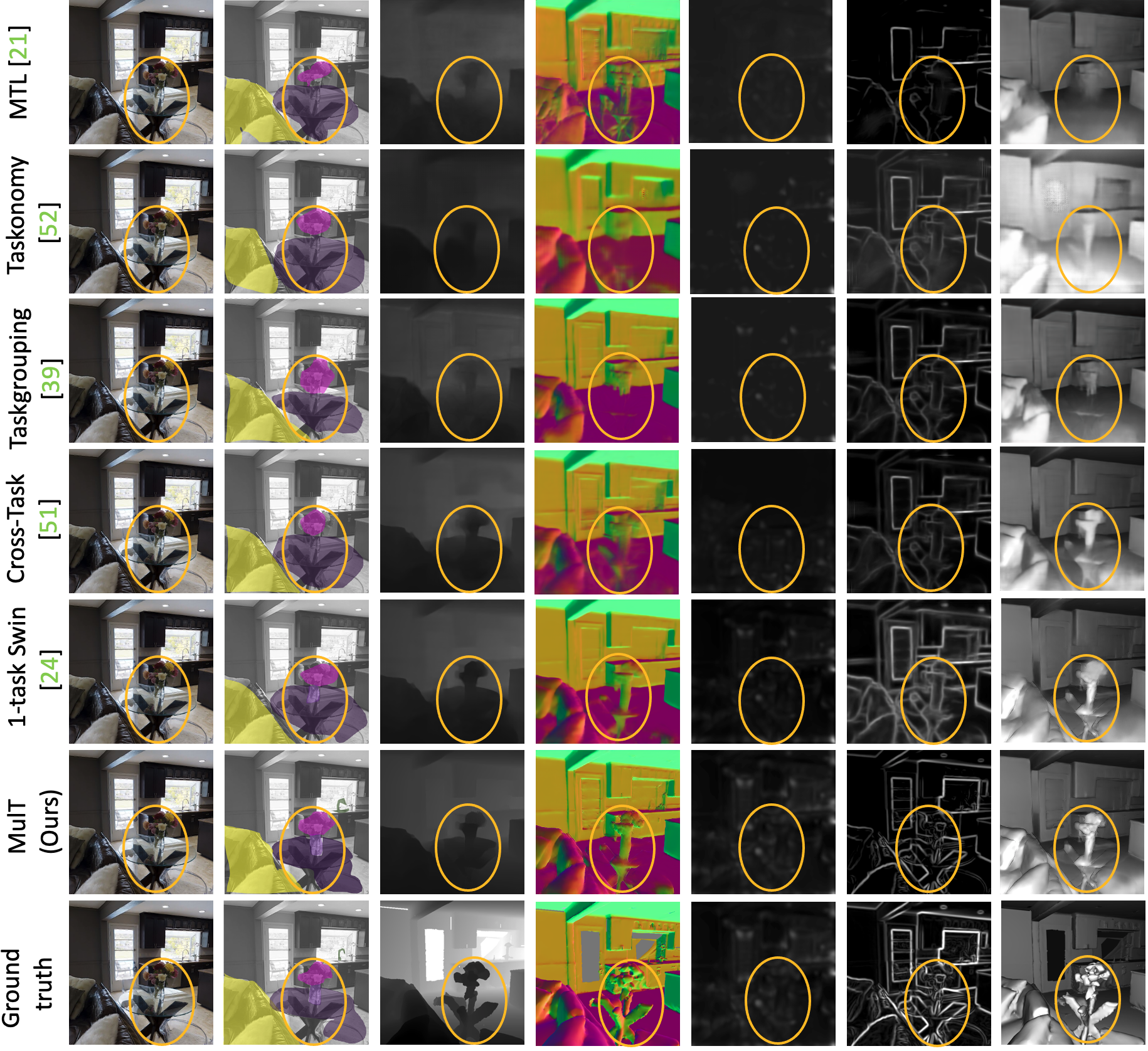}}
%\vspace{-11.3 pt}
\caption{\textbf{Qualitative comparison on the six vision tasks} of the Taskonomy benchmark~\cite{taskonomy2018}. From top to bottom, we show qualitative results using MTL~\cite{kokkinos2016ubernet}, Taskonomy~\cite{taskonomy2018}, Taskgrouping~\cite{standley2019}, Cross-task consistency~\cite{zamir2020consistency}, the single-task dedicated Swin transformer~\cite{swin} and our six-task \textbf{MulT} model. We show, from left to right, the input image, the semantic segmentation results, the depth predictions, the surface normal estimations, the 2D keypoint detections, the 2D edge detections and the reshading results for all the models. All models are jointly trained on the six vision tasks, except for the Swin transformer baseline, which is trained on the independent single tasks. Our MulT model outperforms both the single task Swin baselines and the multitask CNN based baselines. Best seen on screen and zoomed within the yellow circled regions.
}\label{fig:qualitative-result-taskonomybenchmark}\vspace{-10pt}
\end{figure*}
The results in Table~\ref{tb:MulT-pairwise-results} show the relative performance of our MulT model when trained on pairs of tasks and tested on one of the two tasks. We observe that, out of the pairwise-trained multitask models, surface normals help the other vision tasks. However, the performance of normals tends to decrease w.r.t. its single task dedicated model, except when used in conjunction with either depth predition or reshading. Note that the trends we observe are similar to those shown in~\cite{standley2019} for the CNN case. This suggests that transformers follow a similar behavior to that of CNNs in the presence of multiple tasks.
%To overcome, this we further study the effect of reshading on normals. We find that both reshading and depth help to improve the performance of normals. We retrain all the pairwise baseline models keeping thee same settings in~\cite{standley2019} to study the task interdependence. 

In cases of more than two tasks, we observe, as in~\cite{standley2019}, that effectively leveraging between 3 and 6 tasks required increasing the size of the decoder modules. Altogether, reporting results for all possible task combinations requires training $(2^6-1)$ models. Here, we focus on the 6-task case, but provide 3-task, 4-task, and 5-task results in the supplementary material. 
The results of our 6-task MulT model and of the baselines are reported in Table~\ref{tb:MulT-sixstream-results-taskonomy} and Table~\ref{tb:MulT-sixstream-results-replica-nyu} for the Taskonomy test set~\cite{taskonomy2018}, and the Replica~\cite{replica} and NYU~\cite{NYU} dataset, respectively. Our MulT model outperforms the multitask CNN baselines as well as the 1-task CNN and Swin ones. 
Furthermore, as can be verified from the results in the supplementary material, increasing the number of tasks improves the results of our MulT model, e.g., a 6-task network outperforms a 5-task one, which in turn outperforms a 4-task network. 
%Note that as the number of tasks handled by the network increases, the parameter space increases. Therefore, the results for all the Tables are shown with an increased encoder size, as seen in ~\cite{standley2019}.
%An increased encoder size benefits the task performances. In particular, a 6-task network outperforms a 5-task neetwork, which in turn, outperforms a 4-task network. 

\subsection{Qualitative Results}

We qualitatively compare the results of our MulT model with different CNN-based multitask baselines~\cite{kokkinos2016ubernet, taskonomy2018, standley2019, zamir2020consistency}, as well as with the single task dedicated Swin transformer~\cite{swin}. The results in Figure~\ref{fig:qualitative-result-taskonomybenchmark} show the performance of the different networks on all six vision tasks.
%, including, semantic segmentation, depth, surface normals, 2D keypoints, 2D edge and reshading. 
All the multitasking models are jointly trained on the six tasks on the Taskonomy benchmark, and the single task dedicated Swin models are trained on the respective tasks. Our MulT model yields higher-quality predictions than both the single task Swin baselines and the multitask CNN  baselines. We provide additional qualitative results in the supplementary material.

\subsection{Generalization to New Domains}
In this section, we demonstrate how well MulT generalizes to new domains without any fine-tuning,
%and, thereby quantify their robustness, 
and how efficiently MulT can adapt to a new domain by fine-tuning on a small set of training examples from the new domain. To this end, we compare our MulT model and the two baselines of Taskgrouping (TG)~\cite{standley2019} and Cross-task consistency (CT)~\cite{zamir2020consistency} on two new domains, namely, Gaussian-blurred images from Taskonomy~\cite{gaussian} and images from the Cocodoom~\cite{cocodoom} dataset. Note that all the networks were trained on the vanilla Taskonomy dataset~\cite{taskonomy2018}. When fine-tuning the networks, we use either 16 or 128  images from the new domain. The original training data (Taskonomy) is retained during fine-tuning to prevent the networks from forgetting the original domain. 

\begin{figure*}[ht]
\centering
{\includegraphics[ height=6cm, width=13.5cm]{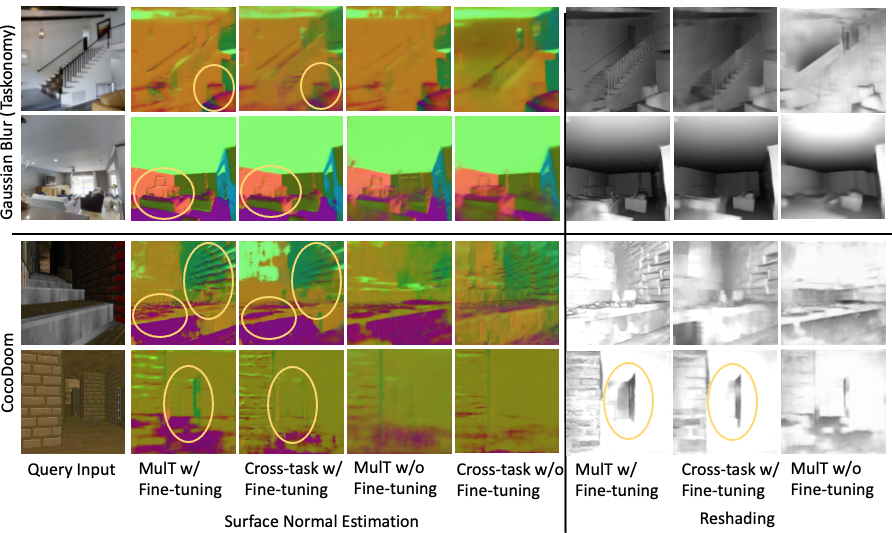}}
%\vspace{-11.3 pt}
\caption{{\textbf{Generalization to new domains.} Our MulT model generalizes better to new domains than the Cross-task~\cite{zamir2020consistency} baseline, both when fine-tuned and not fine-tuned, across the tasks of surface normal prediction and reshading. This shows the benefits of our shared attention module. We test the models on two target domains, Gaussian blur applied to the Taskonomy images~\cite{taskonomy2018} and the out-of-distribution CocoDoom dataset~\cite{cocodoom}. Best viewed on screen and when zoomed in the yellow circled regions.} 
}\label{fig:generalization}\vspace{-10pt}
\end{figure*}

The results in Table~\ref{tb:MulT-generalization-results} and Figure~\ref{fig:generalization} show that our MulT model yields better generalization and adaptation to new domains, both with and without fine-tuning. 
%The better performance of our MulT model is due to both the self attention  and the shared attention. 
These findings confirm the observations made in~\cite{transformer-more-robust-than-cnns} for the single-task scenario. The cross-task consistency~\cite{zamir2020consistency} model shows improved performance in comparison to the Taskgrouping~\cite{standley2019} baseline because of its explicitly enforced consistency constraint, whereas the Taskgrouping model~\cite{standley2019} suffers due to the joint task pairings and the lack of an attention mechanism or an additional constraint. Nevertheless, our MulT model outperforms both these baselines and shows better generalization.

\begin{table}[ht]
\setlength\tabcolsep{1pt}
\centering
\scalebox{0.8}{
\arrayrulecolor{black}
\begin{tabular}{!{\color{white}\vrule}l!{\color{black}\vrule}l!{\color{white}\vrule}c!{\color{white}\vrule}c!{\color{white}\vrule}c!{\color{black}\vrule}l!{\color{white}\vrule}c!{\color{white}\vrule}c}
\hline
\multicolumn{8}{l}{~ ~ ~ ~ ~ ~ ~ ~ ~ ~ ~ ~ ~ ~ ~ ~ ~ ~\textbf{ ~ Generalization to New Domains}}   \\ \hline
                                                    & No. of &\multicolumn{3}{l}{\textbf{~ Error (w/ Fine-tuning)$\downarrow$}} & \multicolumn{3}{l}{ \textbf{Error (w/o Fine-tuning)$\downarrow$}} \\ 
                                        Domains           & images     & MulT & ~ ~CT~\cite{zamir2020consistency} & TG~\cite{standley2019}  & MulT&  ~ ~CT~\cite{zamir2020consistency} & TG~\cite{standley2019} \\ 
\arrayrulecolor{black}\hline
Blur~\cite{gaussian}                                             & 128                                                & \textbf{12.6}                                         &  \underline{17.4}                                           & 21.9                                             & \multirow{2}{*}{\textbf{27.0}}    & \multirow{2}{*}{\underline{46.2}} &       \multirow{2}{*} {55.1}                                                                         \\ 
\small{(Taskonomy)}                                               &  16                                         &  17.5                                             &   22.2                                          & 26.3                                            &              &  &                                                                   \\ 
\hline
CocoDoom                                           & 128                                         &       \textbf{13.3}                                    &   \underline{18.5}                                               & 25.3                                             &  \multirow{2}{*}{\textbf{39.3}}           &\multirow{2}{*}{\underline{ 54.3}} & \multirow{2}{*}{67.7}                                                                        \\ 
                                     ~\cite{cocodoom}    & 16                                         & 20.9                                          & 27.1                                           &  39.9                                                  &    &  &                                   \\
\arrayrulecolor{black}\hline
\end{tabular}}%
\setlength{\abovecaptionskip}{1mm}
\caption{\textbf{Domain generalization on Taskonomy blur data~\cite{gaussian} and CocoDoom~\cite{cocodoom}.} Our MulT model shows better abilities to generalize and adapt to new domains, both with and without fine-tuning. Bold and underlined values show the best and second-best results, respectively. } 
   \label{tb:MulT-generalization-results}%
\vspace{-10pt}
\end{table}%
\vspace{-15pt}
\paragraph{Supplementary Material.} We defer additional discussions and experiments, particularly analyzing the effect of the shared attention in our MulT model and the effect of the network size for different task combinations, as well as additional qualitative results to the supplementary material. We also analyze the number of parameters required by each model and the environmental impact of training such models in the supplementary material.

\section{Conclusion and Limitations}
In this work, we have shown that the transformer framework can be applied to jointly handle multiple tasks within a single end-to-end encoder-decoder framework. Our MulT model simultaneously addresses 6 different vision tasks, learning them in a single training step and
outperforming an independent single task model on each task with a compact set of shared parameters. This allows us to use a single network to handle multiple vision tasks instead of multiple single task networks, thereby reducing the computational cost, for both training and inference. Furthermore, our MulT model outperforms the state-of-the-art CNN-based multitasking models, in terms of both performance in the original domain and generalization/adaptation to new domains.
%, for instance, MulT trained on six vision tasks outperforms the state-of-the-art CNN baselines trained jointly on the same six vision tasks. Moreover, our MulT model can generalize and adapt to unseen domains with lower error, showing its robustness. 

Our current framework nonetheless suffers from some limitations:
\vspace{-15pt}
\paragraph{Data dependency.} Although we validated our findings using various architectures and benchmarks, the results of our approach, as any deep learning one, are in principle data specific. In particular, MulT is a data intensive architecture, and thus when trained on a limited amount of data, it may not achieve the same performance as reported in this work. Note, however, that this is also the case for both single task transformers and the CNN-based multitask baselines.
\vspace{-15pt}
\paragraph{Unpaired Data.} Our current framework, as the CNN-based multitask baselines, requires paired training data. Extending our approach to unlabeled/unpaired data, as in~\cite{zhu2020unpaired, Bhattacharjee_2020_CVPR}, appears feasible and remains open for future work.
\vspace{-15pt}
\paragraph{Modeling efficient attention.} Our current framework makes use of shared attention across the visual tasks. Extending this concept to incorporate local versus global attention, as in~\cite{yang2021focal}, appears feasible and remains open for future work.

Besides addressing these limitations, in the future, we plan to extend our methodology to learning different types of tasks like edge occlusions, principal curvatures and unsupervised segmentation,  and doing zero-shot learning on new tasks. In addition, it would be worthwhile to explore the robustness of large-scale multitask transformers to adversarial tasks, which could become increasingly problematic as the number and variety of tasks grow.

\vspace{2pt}
\textbf{Acknowledgement.} This work was supported in part by the Swiss National Science Foundation via the Sinergia grant CRSII5$-$180359.
%%%%%%%%% REFERENCES
%\newpage
%{\small

%\bibliographystyle{ieee_fullname}
%\bibliography{egbib}
%}

\end{document}

%\pagenumbering{gobble}
\title{MulT: An End-to-End Multitask Learning Transformer \\ \textit{Supplementary}}

\author{Deblina Bhattacharjee, Tong Zhang, Sabine Süsstrunk and Mathieu Salzmann\\
School of Computer and Communication Sciences, EPFL, Switzerland\\
{\tt\small \{deblina.bhattacharjee, tong.zhang, sabine.susstrunk, mathieu.salzmann\}@epfl.ch}
}
\maketitle
We present additional discussions and experiments, particularly the ablation study analyzing the effect of the shared attention in our MulT model, the performances of our 4-task and 5-task networks as well as the ablation study of the effect of the network size on the different task combinations. We also analyze the number of parameters required by each model. We show additional qualitative results comparing the performance of the different models on the Taskonomy~\cite{taskonomy2018} and Replica~\cite{replica} benchmarks. Finally, we discuss the environmental impact of training our model and ways to mitigate it.
The paper is organized as follows:
\begin{itemize}
    \item Section \textbf{\textcolor{red}{1}}: Effect of shared attention 
    \item Section \textbf{\textcolor{red}{2}}: Task combinations
    \item Section \textbf{\textcolor{red}{3}}: Effect of network size
    \item Section \textbf{\textcolor{red}{4}}: Parameter comparison
    \item Section \textbf{\textcolor{red}{5}}: Additional qualitative results
    \item Section \textbf{\textcolor{red}{6}}: Environmental impact
    %\item Section\textbf{\textcolor{red}{7}}: Code
\end{itemize}

\section{Effect of shared attention}
To account for the task dependencies beyond sharing encoder parameters, we develop a shared attention mechanism that integrates the information contained in the encoded features into the decoding stream. Empirically, we have found that the attention from the surface normal task stream benefits our 6-task MulT model and we thus take this task as reference task r, whose attention is shared across the tasks.

In Table~\ref{tb:MulT-shared-attention}, we show the relative performance of our 6-task MulT model with a single-task dedicated Swin transformer baseline~\cite{swin} under two settings. In the first setting the 6-task MulT model is trained \textit{without} the shared attention across the 6 tasks, whereas in the second setting our MulT model is trained \textit{with} the shared attention. The shared attention mechanism benefits the performance of our MulT model, allowing it to learn task inter-dependencies. The models under both the scenarios comprise an increased size of the network. 
\begin{table*}[ht]
\setlength\tabcolsep{3pt}
\centering
\scalebox{0.95}{
\arrayrulecolor{black}
\begin{tabular}{!{\color{white}\vrule}l!{\color{white}\vrule}c!{\color{white}\vrule}c!{\color{white}\vrule}c!{\color{white}\vrule}c!{\color{white}\vrule}c!{\color{white}\vrule}c}
\hline
\multicolumn{7}{l}{~ ~ ~ ~ ~ ~ ~ ~ ~ ~ ~ ~ ~ ~ ~ ~ ~ ~ ~ ~ ~ ~ ~ ~ ~ ~ ~ ~ ~ ~ ~\textbf{ ~ Relative Performance On}}   \\
                                                    & $\mathcal{S}$     & $\mathcal{D}$ & $\mathcal{N}$ & $\mathcal{K}$ & \textit{E} & $\mathcal{R}$  \\ 
\arrayrulecolor{black}\hline
6-task MulT w/o shared attention                                              & \underline{+15.0\%}                                               & \underline{+8.13\% }                                         & \underline{+6.92\% }                                           & \underline{+42.9\% }                                             & \underline{+81.3\% }                                        & \underline{+14.8\% }                                             \\ 
6-task MulT w/ shared attention                                               & \textbf{+19.7\% }                                         & \textbf{+10.2\% }                                                & \textbf{+8.72\%}                                            & \textbf{+94.75\% }                                             & \textbf{+88.8\% }                                        & \textbf{+16.4\% }                                             \\ 
\arrayrulecolor{black}\hline
\end{tabular}}%
\setlength{\abovecaptionskip}{1mm}
\caption{\textbf{Effect of shared attention on our MulT model.} We show the relative performance of our 6-task MulT model with a single-task dedicated Swin transformer baseline~\cite{swin} under two settings- \textit{without} and \textit{with} the shared attention mechanism. Note that under both the settings, our MulT model comprises the increase network size. We show, the relative performance percentage for each task evaluated by taking the percentage increase or decrease w.r.t. the single-task dedicated Swin transformer baseline~\cite{swin}. The shared attention mechanism benefits the performance of our MulT model, allowing it to learn task inter-dependencies. The results here are reported on the Taskonomy test set. Bold and underlined values show the best and second-best results, respectively. }
   \label{tb:MulT-shared-attention}%
\end{table*}%

\paragraph{Feature fusion method.}
We further explore the different feature fusion methods such as concatenation and cross-attention. Concatenating all the features does not benefit our network to learn task interdependencies, as observed in our preliminary experiments, and was thus not reported. Our method \emph{is} a learnable fusion strategy, using a learnable shared attention (SA) mechanism. We also tried the cross-attention (CA) mechanism
from CrossVit~\cite{crossvit}, but it did not beat our SA mechanism in
the given multitask setting, as seen in Table~\ref{tb:MulT-cross-attention} 
\vspace{-12pt}
\begin{table}[ht]
\setlength\tabcolsep{3pt}
\centering
\scalebox{0.75}{
\arrayrulecolor{black}
\begin{tabular}{!{\color{white}\vrule}l!{\color{white}\vrule}c!{\color{white}\vrule}c!{\color{white}\vrule}c!{\color{white}\vrule}c!{\color{white}\vrule}c!{\color{white}\vrule}c}
\hline
\multicolumn{7}{l}{ ~\textbf{ ~ Relative Performance for 6 task MulT vs 1-task SWIN on Taskonomy}}   \\
                                                    & $\mathcal{S}$     & $\mathcal{D}$ & $\mathcal{N}$ & $\mathcal{K}$ & \textit{E} & $\mathcal{R}$  \\ 
\arrayrulecolor{black}\hline
MulT w/ CA                                              & +1.06\%                                              & +5.11\%                                          & -3.33\%                                            & +13.3\%                                              & +25.9\%                                       & +0.06\%                                              \\ 
MulT w/ SA                                               & \textbf{+19.7\% }                                         & \textbf{+10.2\% }                                                & \textbf{+8.72\%}                                            & \textbf{+94.75\% }                                             & \textbf{+88.8\% }                                        & \textbf{+16.4\% }                                             \\ 
\arrayrulecolor{black}\hline
\end{tabular}}%
\setlength{\abovecaptionskip}{1mm}
\caption{\textbf{Quantitative comparison of training the 6-task MulT model on the Taskonomy benchmark~\cite{taskonomy2018} with Cross attention (CA)~\cite{crossvit} and our proposed shared attention (SA).} Our shared attention mechanism benefits MulT where it consistently outperforms the MulT with CA method. Bold values show the best results.  } 
   \label{tb:MulT-cross-attention}%
\end{table}%
\section{Task combinations}
We now show the effect of different task combinations on the relative performance of each task. From our experiments in Table~\ref{tb:task-combinations-network-size}, we observe that the performance of 2D keypoints, 2D edges and segmentation benefits from the inclusion of other tasks like surface normal estimation, depth and reshading. In particular, surface normal estimation is the most beneficial task for the other tasks. For instance, any task with the combination of surface normal estimation, leverages the surface statistics to improve its performance. 

We also observe that increasing the number of tasks improves the results of our MulT model, e.g., a 6-task network outperforms a 5-task one, which in turn outperforms a 4-task network. Note all the models in Table~\ref{tb:task-combinations-network-size} are trained with shared attention to learn task inter-dependencies.

\begin{table*}[ht]
\setlength\tabcolsep{3pt}
\centering
\scalebox{0.75}{
\arrayrulecolor{black}
\begin{tabular}{!{\color{white}\vrule}l!{\color{black}\vrule}l!{\color{black}\vrule}l!{\color{white}\vrule}c!{\color{white}\vrule}c!{\color{white}\vrule}c!{\color{white}\vrule}c!{\color{white}\vrule}c!{\color{black}\vrule}l!{\color{white}\vrule}c!{\color{white}\vrule}c!{\color{white}\vrule}c!{\color{white}\vrule}c!{\color{white}\vrule}c}
\hline
\multicolumn{14}{l}{~ ~ ~ ~ ~ ~ ~ ~ ~ ~ ~ ~ ~ ~ ~ ~ ~ ~  ~ ~ ~ ~ ~ ~ ~ ~ ~ ~ ~ ~ ~ ~ ~\textbf{ ~ Effect of the network size on different task combinations for Taskonomy test~\cite{taskonomy2018}}}   \\
\hline
& &\multicolumn{6}{l}{~ ~ ~ ~ ~ ~ ~ ~ ~ ~ ~ ~ ~ ~ ~ ~\textbf{ w/ increased network size}}& \multicolumn{6}{l}{~ ~ ~ ~ ~ ~ ~ ~ ~ ~ ~ ~ ~\textbf{ w/o increased network size}} \\
                                            No. of
                                            Tasks      & Trained on  & $\mathcal{S}$     & $\mathcal{D}$ & $\mathcal{N}$ & $\mathcal{K}$ & \textit{E} & $\mathcal{R}$ & $\mathcal{S}$     & $\mathcal{D}$ & $\mathcal{N}$ & $\mathcal{K}$ & \textit{E} & $\mathcal{R}$  \\ 
\arrayrulecolor{black}\hline
\multirow{15}{*}{4-task MulT}   &  $\mathcal{S}+\mathcal{D}+\mathcal{N}+\mathcal{K}$                                 & +13.8\%                                                & +8.36\%                                           & +6.91\%                                            &  +82.2\%                & -                                        & -   &+7.84\% &+6.95\% &+5.07\% &+75.4\% &- & -                                         \\ 
&$\mathcal{S}+\mathcal{D}+\mathcal{N}+\textit{E}$                                                  &+14.0\%                                                & +8.38\%                                           & +7.05\%                                            & -               & +74.9\%                                         &-   &+8.08\% &+7.11\% &+5.10\% &- &+63.3\% &-                                                      \\ 
&$\mathcal{S}+\mathcal{D}+\mathcal{N}+\mathcal{R}$                                                  & +14.2\%                                                & +8.55\%                                           & +7.17\%                                            & -               & -                                        & +9.13\%   &+8.11\% &+7.20\% &+5.33\% &- &- & +6.77\% 
               \\ 
&$\mathcal{S}+\mathcal{D}+\mathcal{K}+\textit{E}$                                                   & +13.5\%                                                & +8.08\%                                           & -                                           &  +73.0\%                & +74.6\%                                         & -   &+7.41\% &+6.84\% &- &+67.7\% &+62.7\% & -
               \\ 
&$\mathcal{S}+\mathcal{D}+\mathcal{K}+\mathcal{R}$                                                   & +14.0\%                                                & +8.22\%                                           & -                                            &  +72.4\%                & -                                         & +8.91\%   &+8.03\% &+6.95\% &- &+66.2\% &- & +6.39\% 
               \\ 
&$\mathcal{S}+\mathcal{D}+\textit{E}+\mathcal{R}$                                                   & +14.3\%                                                & +8.30\%                                           & -                                            &  -                & +73.1\%                                         & +9.04\%   &+8.22\% &+6.98\% &- &- &+61.5\% & +6.73\% \\
&$\mathcal{S}+\mathcal{N}+\textit{E}+\mathcal{R}$                                                   & +15.0\%                                                & -                                           & +7.13\%                                            & -                & +73.9\%                                         & +9.17\%   &+8.80\% &- &+5.28\% &- &+61.8\% & +6.80\% \\
&$\mathcal{S}+\mathcal{N}+\mathcal{K}+\mathcal{R}$                                                   & +14.9\%                                                & -                                           & +7.01\%                                            &  +87.5\%                & -                                         & +8.99\%   &+8.61\% &- &+5.12\% &+79.0\% &- & +6.45\% \\
&$\mathcal{S}+\mathcal{N}+\mathcal{K}+\textit{E}$                                                   & +14.7\%                                                & -                                          & +6.89\%                                            &  +88.4\%                & +75.4\%                                         & -   &+8.55\% &- &+5.05\% &+79.7\% &+66.9\% & - \\
&$\mathcal{S}+\mathcal{K}+\textit{E}+\mathcal{R}$                                                   & +13.7\%                                                & -                                           & -                                           &  +73.5\%                & +74.5\%                                         & +8.97\%   &+7.72\% &- &- &+68.9\% &+62.5\% & +6.42\% \\
&$\mathcal{D}+\mathcal{K}+\textit{E}+\mathcal{R}$                                                   & -                                                & +7.91\%                                           & -                                            &  +73.3\%                & +74.8\%                                         & +9.88\%   &- &+6.63\% &- &+68.4\% &+63.0\% & +7.00\% \\
&$\mathcal{D}+\mathcal{N}+\mathcal{K}+\mathcal{R}$                                                   &-                                                & +8.44\%                                           & +7.20\%                                            &  +87.0\%                & -                                        & +10.3\%   &- &+7.15\% &+5.40\% &+78.8\% &- & +7.33\% \\
&$\mathcal{D}+\mathcal{N}+\textit{E}+\mathcal{R}$                                                   & -                                               & +8.63\%                                           & +7.25\%                                            &  -               & +75.5\%                                         & +11.1\%   &- &+7.29\% &+5.49\% &-&+66.8\% & +8.12\% \\
&$\mathcal{D}+\mathcal{N}+\mathcal{K}+\textit{E}$                                                   & -                                                & +8.40\%                                           & +7.10\%                                            &  +87.2\%                & +75.8\%                                         & -   &- &+7.12\% &+5.20\% &+79.2\% &+67.0\% & - \\
&$\mathcal{N}+\mathcal{K}+\textit{E}+\mathcal{R}$                                                   & -                                                & -                                           & +7.12\%                                            & +88.8\%                & +75.0\%                                         & +10.6\%   &- &- &+5.27\% &+80.1\% &+66.1\% & +7.74\% \\
\hline
\multirow{6}{*}{5-task MulT}   &  $\mathcal{S}+\mathcal{D}+\mathcal{N}+\mathcal{K}+\textit{E}$                                         &+17.2\%                                                & +9.07\%                                           &+8.11 \%                                            & +92.5 \%                & +82.6\%                                         & -   &+11.6\% &+7.75\% &+6.16\% &+89.9\% &+72.5\% &-                                          \\ 
&$\mathcal{S}+\mathcal{D}+\mathcal{N}+\mathcal{K}+\mathcal{R}$                                                  & +17.7\%                                                & +9.10\%                                           & +7.59\%                                            &  +92.0\%                &-                                         & +12.9\%   &+12.0\% &+7.91\% &+5.94\% &+89.5\% &- &+10.0 \%                                                      \\ 
&$\mathcal{S}+\mathcal{D}+\mathcal{N}+\textit{E}+\mathcal{R}$                                                  & +16.9\%                                                & +9.22\%                                           & \underline{+8.26\%}                                            & -                & \underline{+82.9\%}                                         & +12.7\%   &+10.8\% &+8.08\% &\underline{+6.47\%} &- &\underline{+72.9\%} &+9.71\% 
               \\ 
&$\mathcal{S}+\mathcal{D}+\mathcal{K}+\textit{E}+\mathcal{R}$                                                   & +15.1\%                                                & +8.86\%      &-                                     & +75.0\%                                            &  +78.8\%                & +10.2\%                                         & +9.10\%   &+7.47\% &- &+70.7\% &+67.7\% &+7.80\% \\
&$\mathcal{S}+\mathcal{N}+\mathcal{K}+\textit{E}+\mathcal{R}$                                                   & \underline{+18.3\%}                                            &-                                           & +7.33\%                                            &  \underline{+94.1\%}                & +82.2\%                                         & +13.0\%   &\underline{+12.5\%} &- &+5.55\% &\underline{+91.9\%} &+72.2\% & +10.3\% 
               \\ 
&$\mathcal{D}+\mathcal{N}+\mathcal{K}+\textit{E}+\mathcal{R}$                                                   & -                                                & +\underline{9.77\%}                                         & +8.06\%                                            & +93.9\%                &+82.6\%                                         & \underline{+13.8\% }  &- &\underline{+8.33\%}&+6.11\% &+91.6\% &+72.5\% &\underline{+10.7\% }
               \\ 
\hline
6-task MulT& $\mathcal{S}+\mathcal{D}+\mathcal{N}+\mathcal{K}+\textit{E}+\mathcal{R}$                                                   &\textbf{+19.7\%} &\textbf{+10.2\%} &\textbf{+8.72\%} &\textbf{+94.7\%} &\textbf{+88.8\%}&\textbf{+16.4\%}  &\textbf{+13.8\%}&\textbf{+9.11\%}&\textbf{+6.99\%} &\textbf{+92.5\%} &\textbf{+78.3\%} &\textbf{+12.9\%}   \\                 
\arrayrulecolor{black}\hline
\end{tabular}}%
\setlength{\abovecaptionskip}{1mm}
\caption{\textbf{Quantitative comparison of training different task combinations in our MulT model on the Taskonomy benchmark~\cite{taskonomy2018}.} Increasing the number of tasks improves the results of our MulT models, where a 6-task network outperforms a 5-task one, which in turn outperforms a 4-task network. Note all the models are trained with shared attention to learn task inter-dependencies. The relative performance percentage for each task is evaluated by taking the percentage increase or decrease w.r.t. the single-task swin~\cite{swin} baseline. Bold and underlined values show the best and second-best results, respectively.  } 
   \label{tb:task-combinations-network-size}%
\vspace{-10pt}
\end{table*}%

%%%%%%%%%%%%
\mycomment{
\begin{table*}[ht]
\setlength\tabcolsep{3pt}
\centering
\scalebox{0.75}{
\arrayrulecolor{black}
\begin{tabular}{!{\color{white}\vrule}l!{\color{black}\vrule}l!{\color{black}\vrule}l!{\color{white}\vrule}c!{\color{white}\vrule}c!{\color{white}\vrule}c!{\color{white}\vrule}c!{\color{white}\vrule}c!{\color{black}\vrule}l!{\color{white}\vrule}c!{\color{white}\vrule}c!{\color{white}\vrule}c!{\color{white}\vrule}c!{\color{white}\vrule}c}
\hline
\multicolumn{14}{l}{~ ~ ~ ~ ~ ~ ~ ~ ~ ~ ~ ~ ~ ~ ~ ~ ~ ~  ~ ~ ~ ~ ~ ~ ~ ~ ~ ~ ~ ~ ~ ~ ~\textbf{ ~ Effect of the network size on different task combinations for Taskonomy test~\cite{taskonomy2018}}}   \\
\hline
& &\multicolumn{6}{l}{~ ~ ~ ~ ~ ~ ~ ~ ~ ~ ~ ~ ~ ~ ~ ~\textbf{ w/ increased network size}}& \multicolumn{6}{l}{~ ~ ~ ~ ~ ~ ~ ~ ~ ~ ~ ~ ~\textbf{ w/o increased network size}} \\
                                            No. of
                                            Tasks      & Trained on  & $\mathcal{S}$     & $\mathcal{D}$ & $\mathcal{N}$ & $\mathcal{K}$ & \textit{E} & $\mathcal{R}$ & $\mathcal{S}$     & $\mathcal{D}$ & $\mathcal{N}$ & $\mathcal{K}$ & \textit{E} & $\mathcal{R}$  \\ 
\arrayrulecolor{black}\hline
\multirow{20}{*}{3-task MulT}   &  $\mathcal{S}+\mathcal{D}+\mathcal{N}$                                         & \%                                                & \%                                           & \%                                            &  \%                & \%                                         & \%   &\% &\% &\% &\% &\% & \%                                          \\ 
&$\mathcal{S}+\mathcal{D}+\mathcal{K}$                                                  & \%                                                & \%                                           & \%                                            &  \%                & \%                                         & \%   &\% &\% &\% &\% &\% & \%                                                      \\ 
&$\mathcal{S}+\mathcal{D}+\textit{E}$                                                  & \%                                                & \%                                           & \%                                            &  \%                & \%                                         & \%   &\% &\% &\% &\% &\% & \% 
               \\ 
&$\mathcal{S}+\mathcal{D}+\mathcal{R}$                                                   & \%                                                & \%                                           & \%                                            &  \%                & \%                                         & \%   &\% &\% &\% &\% &\% & \% 
               \\ 
&$\mathcal{S}+\mathcal{N}+\mathcal{K}$                                                   & \%                                                & \%                                           & \%                                            &  \%                & \%                                         & \%   &\% &\% &\% &\% &\% & \% 
               \\ 
&$\mathcal{S}+\mathcal{N}+\textit{E}$                                                   & \%                                                & \%                                           & \%                                            &  \%                & \%                                         & \%   &\% &\% &\% &\% &\% & \% \\
&$\mathcal{S}+\mathcal{N}+\mathcal{R}$                                                   & \%                                                & \%                                           & \%                                            &  \%                & \%                                         & \%   &\% &\% &\% &\% &\% & \% \\
&$\mathcal{S}+\textit{E}+\mathcal{R}$                                                  & \%                                                & \%                                           & \%                                            &  \%                & \%                                         & \%   &\% &\% &\% &\% &\% & \%                                                      \\
&$\mathcal{S}+\mathcal{K}+\textit{E}$                                                   & \%                                                & \%                                           & \%                                            &  \%                & \%                                         & \%   &\% &\% &\% &\% &\% & \% \\
&$\mathcal{S}+\mathcal{K}+\mathcal{R}$                                                   & \%                                                & \%                                           & \%                                            &  \%                & \%                                         & \%   &\% &\% &\% &\% &\% & \% \\
&$\mathcal{D}+\mathcal{N}+\mathcal{K}$                                                   & \%                                                & \%                                           & \%                                            &  \%                & \%                                         & \%   &\% &\% &\% &\% &\% & \% \\
&$\mathcal{D}+\mathcal{N}+\textit{E}$                                                   & \%                                                & \%                                           & \%                                            &  \%                & \%                                         & \%   &\% &\% &\% &\% &\% & \% \\
&$\mathcal{D}+\mathcal{N}+\mathcal{R}$                                                   & \%                                                & \%                                           & \%                                            &  \%                & \%                                         & \%   &\% &\% &\% &\% &\% & \% \\
&$\mathcal{D}+\mathcal{K}+\textit{E}$                                                  & \%                                                & \%                                           & \%                                            &  \%                & \%                                         & \%   &\% &\% &\% &\% &\% & \%                                                      \\ 
&$\mathcal{D}+\mathcal{K}+\mathcal{R}$                                                  & \%                                                & \%                                           & \%                                            &  \%                & \%                                         & \%   &\% &\% &\% &\% &\% & \%                                                      \\ 
&$\mathcal{D}+\textit{E}+\mathcal{R}$                                                  & \%                                                & \%                                           & \%                                            &  \%                & \%                                         & \%   &\% &\% &\% &\% &\% & \%                                                      \\ 
&$\mathcal{N}+\textit{E}+\mathcal{R}$                                                  & \%                                                & \%                                           & \%                                            &  \%                & \%                                         & \%   &\% &\% &\% &\% &\% & \%                                                      \\ 
&$\mathcal{N}+\mathcal{K}+\textit{E}$                                                   & \%                                                & \%                                           & \%                                            &  \%                & \%                                         & \%   &\% &\% &\% &\% &\% & \% \\
&$\mathcal{N}+\mathcal{K}+\mathcal{R}$                                                   & \%                                                & \%                                           & \%                                            &  \%                & \%                                         & \%   &\% &\% &\% &\% &\% & \% \\
&$\mathcal{K}+\textit{E}+\mathcal{R}$                                                   & \%                                                & \%                                           & \%                                            &  \%                & \%                                         & \%   &\% &\% &\% &\% &\% & \% \\
\hline
\multirow{15}{*}{4-task MulT}   &  $\mathcal{S}+\mathcal{D}+\mathcal{N}+\mathcal{K}$                                 & +13.8\%                                                & +8.36\%                                           & +6.91\%                                            &  +82.2\%                & -                                        & -   &+7.84\% &+6.95\% &+5.07\% &+75.4\% &- & -                                         \\ 
&$\mathcal{S}+\mathcal{D}+\mathcal{N}+\textit{E}$                                                  &+14.0\%                                                & +8.38\%                                           & +7.05\%                                            & -               & +74.9\%                                         &-   &+8.08\% &+7.11\% &+5.10\% &- &+63.3\% &-                                                      \\ 
&$\mathcal{S}+\mathcal{D}+\mathcal{N}+\mathcal{R}$                                                  & +14.2\%                                                & +8.55\%                                           & +7.17\%                                            & -               & -                                        & +9.13\%   &+8.11\% &+7.20\% &+5.33\% &- &- & +6.77\% 
               \\ 
&$\mathcal{S}+\mathcal{D}+\mathcal{K}+\textit{E}$                                                   & +13.5\%                                                & +8.08\%                                           & -                                           &  +73.0\%                & +74.6\%                                         & -   &+7.41\% &+6.84\% &- &+67.7\% &+62.7\% & -
               \\ 
&$\mathcal{S}+\mathcal{D}+\mathcal{K}+\mathcal{R}$                                                   & +14.0\%                                                & +8.22\%                                           & -                                            &  +72.4\%                & -                                         & +8.91\%   &+8.03\% &+6.95\% &- &+66.2\% &- & +6.39\% 
               \\ 
&$\mathcal{S}+\mathcal{D}+\textit{E}+\mathcal{R}$                                                   & +14.3\%                                                & +8.30\%                                           & -                                            &  -                & +73.1\%                                         & +9.04\%   &+8.22\% &+6.98\% &- &- &+61.5\% & +6.73\% \\
&$\mathcal{S}+\mathcal{N}+\textit{E}+\mathcal{R}$                                                   & +15.0\%                                                & -                                           & +7.13\%                                            & -                & +73.9\%                                         & +9.17\%   &+8.80\% &- &+5.28\% &- &+61.8\% & +6.80\% \\
&$\mathcal{S}+\mathcal{N}+\mathcal{K}+\mathcal{R}$                                                   & +14.9\%                                                & -                                           & +7.01\%                                            &  +87.5\%                & -                                         & +8.99\%   &+8.61\% &- &+5.12\% &+79.0\% &- & +6.45\% \\
&$\mathcal{S}+\mathcal{N}+\mathcal{K}+\textit{E}$                                                   & +14.7\%                                                & -                                          & +6.89\%                                            &  +88.4\%                & +75.4\%                                         & -   &+8.55\% &- &+5.05\% &+79.7\% &+66.9\% & - \\
&$\mathcal{S}+\mathcal{K}+\textit{E}+\mathcal{R}$                                                   & +13.7\%                                                & -                                           & -                                           &  +73.5\%                & +74.5\%                                         & +8.97\%   &+7.72\% &- &- &+68.9\% &+62.5\% & +6.42\% \\
&$\mathcal{D}+\mathcal{K}+\textit{E}+\mathcal{R}$                                                   & -                                                & +7.91\%                                           & -                                            &  +73.3\%                & +74.8\%                                         & +9.88\%   &- &+6.63\% &- &+68.4\% &+63.0\% & +7.00\% \\
&$\mathcal{D}+\mathcal{N}+\mathcal{K}+\mathcal{R}$                                                   &-                                                & +8.44\%                                           & +7.20\%                                            &  +87.0\%                & -                                        & +10.3\%   &- &+7.15\% &+5.40\% &+78.8\% &- & +7.33\% \\
&$\mathcal{D}+\mathcal{N}+\textit{E}+\mathcal{R}$                                                   & -                                               & +8.63\%                                           & +7.25\%                                            &  -               & +75.5\%                                         & +11.1\%   &- &+7.29\% &+5.49\% &-&+66.8\% & +8.12\% \\
&$\mathcal{D}+\mathcal{N}+\mathcal{K}+\textit{E}$                                                   & -                                                & +8.40\%                                           & +7.10\%                                            &  +87.2\%                & +75.8\%                                         & -   &- &+7.12\% &+5.20\% &+79.2\% &+67.0\% & - \\
&$\mathcal{N}+\mathcal{K}+\textit{E}+\mathcal{R}$                                                   & -                                                & -                                           & +7.12\%                                            & +88.8\%                & +75.0\%                                         & +10.6\%   &- &- &+5.27\% &+80.1\% &+66.1\% & +7.74\% \\

\hline
\multirow{6}{*}{5-task MulT}   &  $\mathcal{S}+\mathcal{D}+\mathcal{N}+\mathcal{K}+\textit{E}$                                         &+17.2\%                                                & +9.07\%                                           &+8.11 \%                                            & +92.5 \%                & +82.6\%                                         & -   &+11.6\% &+7.75\% &+6.16\% &+89.9\% &+72.5\% &-                                          \\ 
&$\mathcal{S}+\mathcal{D}+\mathcal{N}+\mathcal{K}+\mathcal{R}$                                                  & +17.7\%                                                & +9.10\%                                           & +7.59\%                                            &  +92.0\%                &-                                         & +12.9\%   &+12.0\% &+7.91\% &+5.94\% &+89.5\% &- &+10.0 \%                                                      \\ 
&$\mathcal{S}+\mathcal{D}+\mathcal{N}+\textit{E}+\mathcal{R}$                                                  & +16.9\%                                                & +9.22\%                                           & \underline{+8.26\%}                                            & -                & \underline{+82.9\%}                                         & +12.7\%   &+10.8\% &+8.08\% &\underline{+6.47\%} &- &\underline{+72.9\%} &+9.71\% 
               \\ 
&$\mathcal{S}+\mathcal{D}+\mathcal{K}+\textit{E}+\mathcal{R}$                                                   & +15.1\%                                                & +8.86\%      &-                                     & +75.0\%                                            &  +78.8\%                & +10.2\%                                         & +9.10\%   &+7.47\% &- &+70.7\% &+67.7\% &+7.80\% \\
&$\mathcal{S}+\mathcal{N}+\mathcal{K}+\textit{E}+\mathcal{R}$                                                   & \underline{+18.3\%}                                            &-                                           & +7.33\%                                            &  \underline{+94.1\%}                & +82.2\%                                         & +13.0\%   &\underline{+12.5\%} &- &+5.55\% &\underline{+91.9\%} &+72.2\% & +10.3\% 
               \\ 
&$\mathcal{D}+\mathcal{N}+\mathcal{K}+\textit{E}+\mathcal{R}$                                                   & -                                                & +\underline{9.77\%}                                         & +8.06\%                                            & +93.9\%                &+82.6\%                                         & \underline{+13.8\% }  &- &\underline{+8.33\%}&+6.11\% &+91.6\% &+72.5\% &\underline{+10.7\% }
               \\ 
\hline
6-task MulT& $\mathcal{S}+\mathcal{D}+\mathcal{N}+\mathcal{K}+\textit{E}+\mathcal{R}$                                                   &\textbf{+19.7\%} &\textbf{+10.2\%} &\textbf{+8.72\%} &\textbf{+94.7\%} &\textbf{+88.8\%}&\textbf{+16.4\%}  &\textbf{+13.8\%}&\textbf{+9.11\%}&\textbf{+6.99\%} &\textbf{+92.5\%} &\textbf{+78.3\%} &\textbf{+12.9\%}   \\                 
\arrayrulecolor{black}\hline
\end{tabular}}%
\setlength{\abovecaptionskip}{1mm}
\caption{\textbf{Quantitative comparison of training different task combinations in our MulT model on the Taskonomy benchmark~\cite{taskonomy2018}.} Increasing the number of tasks improves the results of our MulT models, where a 6-task network outperforms a 5-task one, which in turn outperforms a 4-task network. Note all the models are trained with shared attention to learn task inter-dependencies. The relative performance percentage for each task is evaluated by taking the percentage increase or decrease w.r.t. the single-task swin~\cite{swin} baseline. Bold and underlined values show the best and second-best results, respectively.  } 
   \label{tb:task-combinations-network-size-updated}%
\vspace{-10pt}
\end{table*}%
}

\vspace{-5pt}
\section{Effect of network size}
As more number of tasks are added to our MulT model, we observe, as in~\cite{standley2019}, that effectively leveraging between 3 and 6 tasks required increasing the size of the network modules. Altogether, reporting results for all possible task combinations requires training $(2^6-1)$ models.
 We see that improving the network size has significant effect on the relative performance of the different tasks. We quantitatively evaluate all the task combinations in the %3-task,
 4-task, 5-task and 6-task settings; with and without an increase in the network size. For the normal network size, we use swin-T as the backbone~\cite{swin} containing ${(2,2,6,2)}$ transformer blocks in the respective stages of the encoder, whereas for the increased network we use swin-L~\cite{swin} as the backbone containing ${(2,2,18,2)}$ transformer blocks in the respective stages of the encoder. The increase in the number of transformer blocks in the third stage of the encoder in the swin-L backbone helps to accommodate the increased number of tasks. Our best performing MulT model comprises the increased network size and shared attention.
 
 In Table~\ref{tb:task-combinations-network-size}, we observe that increasing the number of tasks improves the results of our MulT model, where a 6-task network outperforms a 5-task one, which in turn outperforms a 4-task network. Note all the models in Table~\ref{tb:task-combinations-network-size} are trained with shared attention to learn task inter-dependencies.
 
\section{Parameter comparison}
We show the number of parameters learnt by our 6-task MulT model \textit{without} an increased network size and compare it to the number of parameters learnt by the multitasking Resnet50 baseline and the single dedicated Swin-Tiny (Swin-T) baseline. Further, we show the number of parameters learnt by our 6-task MulT model \textit{with} an increased network size and compare it to the number of parameters learnt by the multitasking Resnet152 baseline and the single dedicated Swin-Large (Swin-L) baseline. We see that our MulT model, both without and with an increased network size, is more parameter efficient than the 1-task dedicated Swin-T and Swin-L models, respectively. Note that the number of parameters and the inference time of six 1-task Swin-T models and six 1-task Swin-L models are added to get the total number of parameters and the total inference time for all the six tasks. Our MulT model learns more number of parameters than the multitasking CNN baselines~\cite{resnet} but infers the final predictions across the six tasks in comparable time.
\begin{table}[ht]
\setlength\tabcolsep{3pt}
\centering
\scalebox{0.8}{
\arrayrulecolor{black}
\begin{tabular}{!{\color{white}\vrule}l!{\color{white}\vrule}c!{\color{white}\vrule}c}
\hline
\multicolumn{3}{l}{~ ~ ~ ~ ~ ~ ~ ~ ~ ~ ~ ~ ~ ~ ~ ~ ~\textbf{ ~ Parameter Comparison}}   \\
                                              Model      & No. of Params (M)     & Inference time (ms) \\ 
\arrayrulecolor{black}\hline
Multitasking Resnet50~\cite{resnet}                                              & 153.6                                               & 12                                       \\ 
six 1-task Swin-T~\cite{swin}                                              & 344                                               & 90                                       \\ 
MulT w/o increased network                                        & 231                                               & 13                                       \\ 
\hline
Multitasking Resnet152~\cite{resnet}                                              & 361.2                                               & 27                                      \\ 
six 1-task Swin-L~\cite{swin}                                                & 728                                               & 198                                       \\ 
MulT w/ increased network                                              & 545                                             & 29                                      \\ 
\arrayrulecolor{black}\hline
\end{tabular}}%
\setlength{\abovecaptionskip}{1mm}
\caption{\textbf{Parameter comparison of our 6-task MulT model with the baselines.} We see that our MulT model, both without and with an increased network size, is more parameter efficient than the 1-task dedicated Swin-T and Swin-L models, respectively. Note that the number of parameters and the inference time of six 1-task Swin-T models and six 1-task Swin-L models are added to get the total number of parameters and the total inference time for all the six tasks. Further, our MulT model learns more number of parameters than the multitasking CNN baselines~\cite{resnet} but infers the final predictions across the six tasks in comparable time.}
\label{tb:parameter-comparison}%
\vspace{-10pt}
\end{table}%

\section{Additional qualitative results}
\begin{figure*}[ht]
\centering
{\includegraphics[ height= 9.5cm, width=14.3cm]{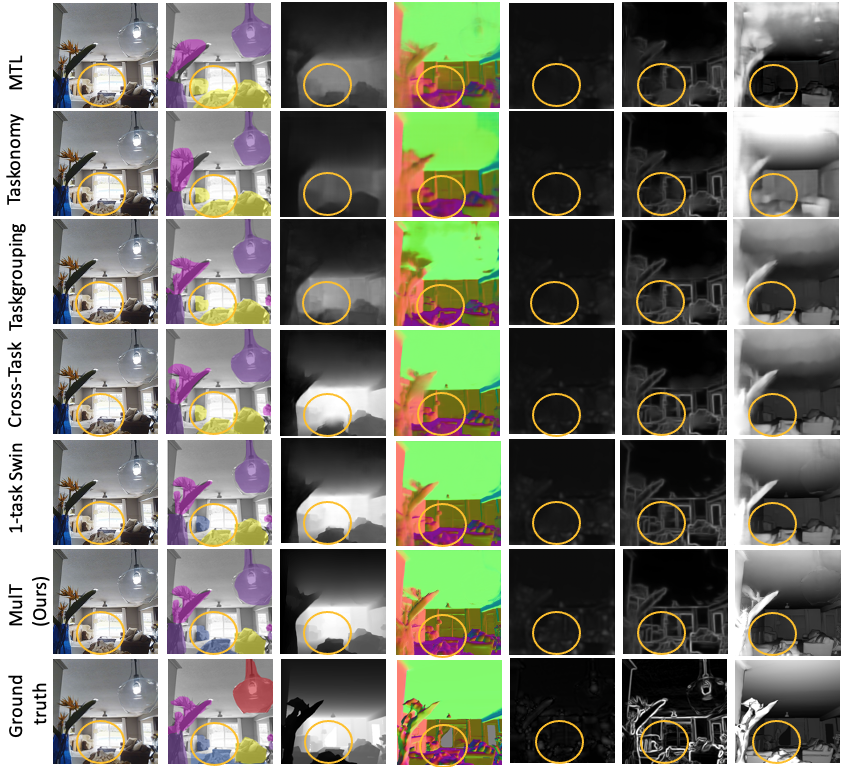}}
\vspace{13pt}
{\includegraphics[ height= 9.5cm, width=14.3cm]{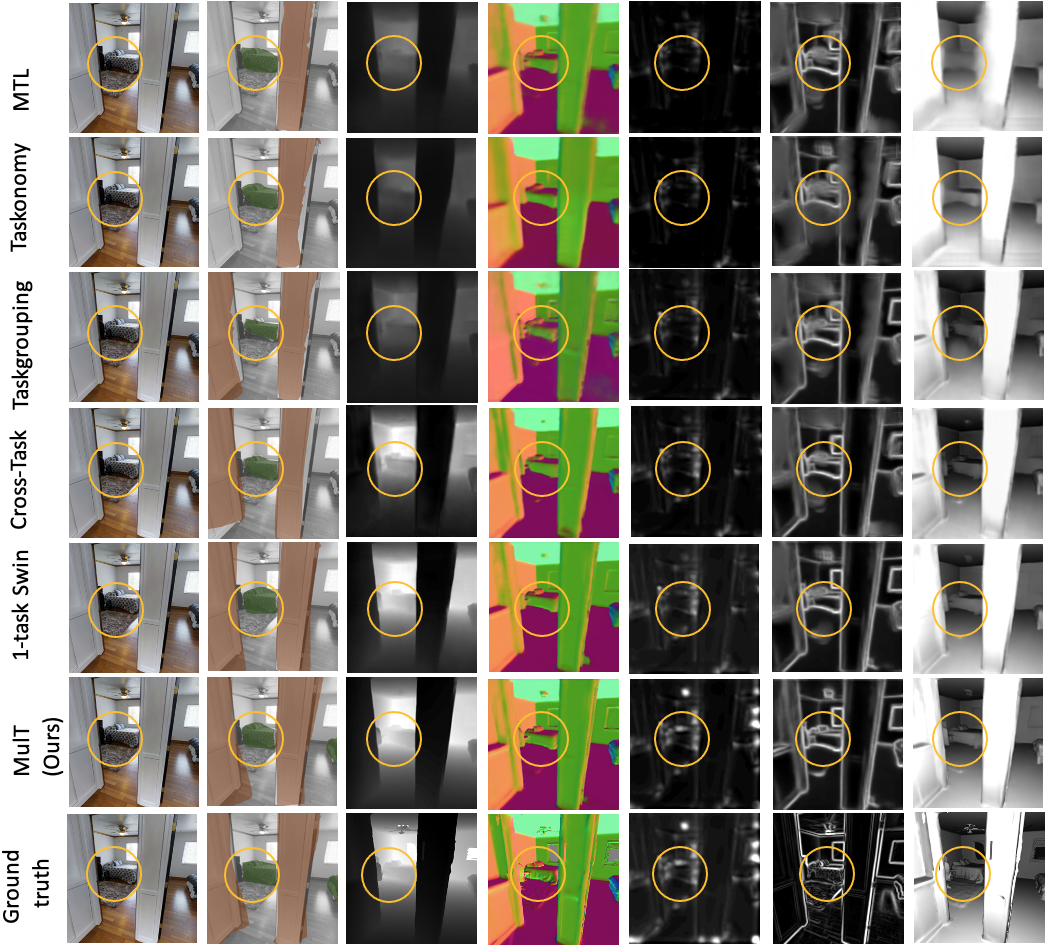}}
\caption{\textbf{Qualitative comparison on the six vision tasks} of the Taskonomy benchmark~\cite{taskonomy2018}. From top to bottom, we show qualitative results using MTL~\cite{kokkinos2016ubernet}, Taskonomy~\cite{taskonomy2018}, Taskgrouping~\cite{standley2019}, Cross-task consistency~\cite{zamir2020consistency}, the single-task dedicated Swin transformer~\cite{swin} and our six-task \textbf{MulT} model. We show, from left to right, the input image, the semantic segmentation results, the depth predictions, the surface normal estimations, the 2D keypoint detections, the 2D edge detections and the reshading results for all the models. All models are jointly trained on the six vision tasks, except for the Swin transformer baseline, which is trained on the independent single tasks. Our MulT model outperforms both the single task Swin baselines and the multitask CNN based baselines. Best seen on screen and zoomed within the yellow circled regions.
}\label{fig:qualitative-result-taskonomybenchmark1}\vspace{-10pt}
\end{figure*}

\begin{figure*}[ht]
\centering
{\includegraphics[ height= 8.5cm, width=10.3cm]{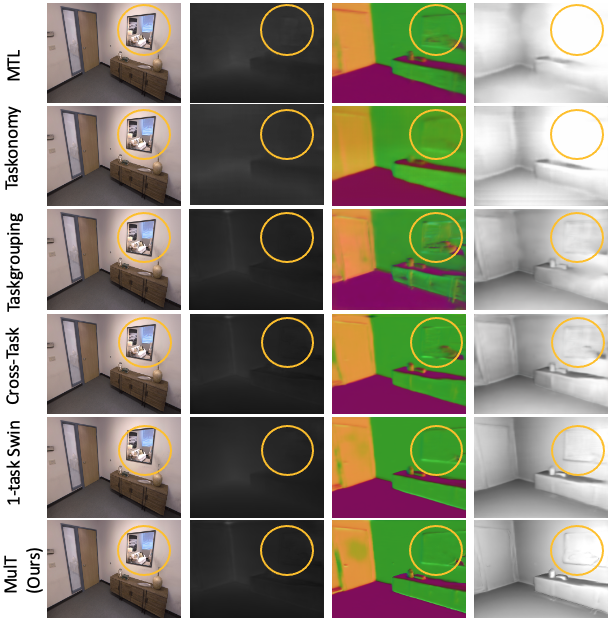}}
\vspace{13pt}
{\includegraphics[ height= 9.7cm, width=10.3cm]{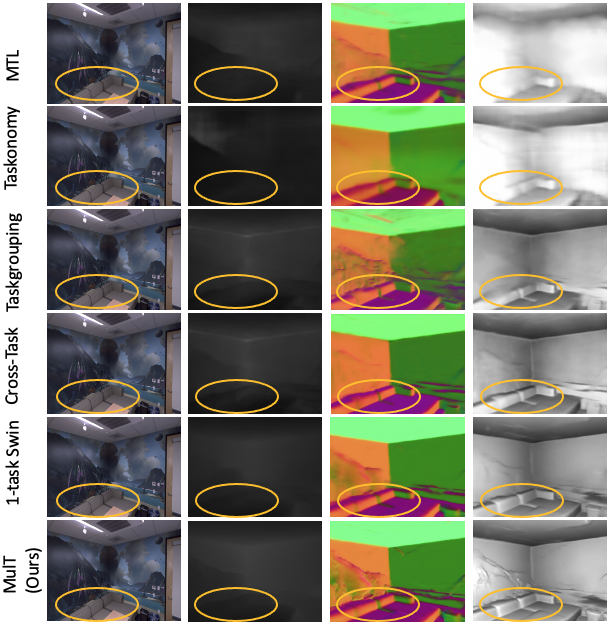}}
\caption{\textbf{Qualitative comparison on the three vision tasks} of the Replica benchmark~\cite{replica}. From top to bottom, we show qualitative results using MTL~\cite{kokkinos2016ubernet}, Taskonomy~\cite{taskonomy2018}, Taskgrouping~\cite{standley2019}, Cross-task consistency~\cite{zamir2020consistency}, the single-task dedicated Swin transformer~\cite{swin} and our six-task \textbf{MulT} model. We show, from left to right, the input image, the depth predictions, the surface normal estimations and the reshading results for all the models. All models are jointly trained on the \textit{six} vision tasks of the Taskonomy benchmark and are then fine-tuned to the Replica official training set, except for the Swin transformer baseline, which is trained on the independent \textit{single} tasks. Our MulT model outperforms both the single task Swin baselines and the multitask CNN based baselines. Best seen on screen and zoomed within the yellow circled regions.
}\label{fig:qualitative-result-replica}\vspace{-10pt}
\end{figure*}

We qualitatively compare the results of our MulT model with different CNN-based multitask baselines~\cite{kokkinos2016ubernet, taskonomy2018, standley2019, zamir2020consistency}, as well as with the single task dedicated Swin transformer~\cite{swin}. The results in Figure~\ref{fig:qualitative-result-taskonomybenchmark1} and Figure~\ref{fig:qualitative-result-replica} show the performance of the different networks across multiple vision tasks on the Taskonomy benchmark~\cite{taskonomy2018} and Replica test set~\cite{replica}, respectively.
All the multitasking models are jointly trained on the six tasks on the Taskonomy benchmark, and the single task dedicated Swin models are trained on the respective tasks. Our MulT model yields higher-quality predictions than both the single task Swin baselines and the multitask CNN  baselines. 
%%%%%%%%%%%%%%
\section{Environmental impact}
Models consume power both during training as well as during inference. However, a bigger source of energy consumption today comes from after the models are deployed, i.e. during the inference stage~\cite{consumption-study}. Nvidia estimated that in 2019, 80–90\% of the cost of a model is in the inference. To worsen this, machine learning practitioners waste a ton of resources on redundant training~\cite{consumption-study}.
By being a multitask framework, our MulT model helps to reduce the power consumption during inference unlike the single task baselines that need to be run multiple times to achieve the predictions on the different tasks. A shown in Table~\ref{tb:parameter-comparison}, our MulT model requires less inference time than the single task transformer baselines while reporting better performance.
Nonetheless, running our MulT model in the cloud takes 21 hours to train on 32 Nvidia V100-SXM2-32GB GPUs, where a single GPU emits 3.11kg of $CO_2$ with a $CO_2$ offset of 1.55kg~\cite{mlco2-calculator}. This is equivalent to 12.5 kilometers driven by an average car~\cite{lacoste2019quantifying}.
To mitigate, the carbon footprint of training our model we have reputable carbon offsets as well as follow a centralised cloud infrastructure with sustainable power supplies. Furthermore, by employing an efficient shared attention mechanism as~\cite{yang2021focal}, that operates in linear time, we can extend our mitigation efforts and reduce the overall hours of GPU computation.
%\section{Code}

%%%%%%%%% REFERENCES

%{\small
%\bibliographystyle{ieee_fullname}
%\bibliography{egbib}
%}